%% file: main.tex
\documentclass[11pt]{article}

\input{math_commands.tex}

\usepackage[preprint]{acl}

\usepackage{times}
\usepackage{latexsym}

\usepackage[T1]{fontenc}

\usepackage[utf8]{inputenc}

\usepackage{microtype}
\usepackage{tabularx}

\usepackage{inconsolata}
\usepackage{adjustbox}
\usepackage{graphicx}
\usepackage{subcaption}
\usepackage{booktabs}
\usepackage{multicol}
\usepackage{multirow}
\usepackage[most]{tcolorbox}   
\usepackage{xcolor}   
\usepackage{enumitem}
\newtcolorbox{chatbubble}[2]{
  enhanced, breakable, sharp corners,
  colback=#1!6,        
  colframe=#1!60!black,
  boxrule=.45pt,
  boxsep=3pt,
  left=4pt,right=4pt,top=3pt,bottom=2pt,
  title      = {#2},
  fonttitle  = \bfseries\color{white},
  boxed title style = {
    colback=#1!70!black,
    frame hidden,
    sharp corners,
    left=4pt,right=4pt,top=1pt,bottom=1pt,
  },
  attach boxed title to top left = {
    xshift=-.45pt,
    yshift=-\tcboxedtitleheight 
  },
}

\title{PRISM: A Unified Framework for Post-Training LLMs Without Verifiable Rewards}

\author{
 \textbf{Mukesh Ghimire\textsuperscript{1$\dagger$}},
 \textbf{Aosong Feng\textsuperscript{2}},
 \textbf{Liwen You\textsuperscript{2}},
 \textbf{Youzhi Luo\textsuperscript{2}},
 \textbf{Fang Liu\textsuperscript{2}},
 \textbf{Xuan Zhu\textsuperscript{2}}
\\
\\
 \textsuperscript{1}Arizona State University,
 \textsuperscript{2}Amazon Web Services
\\
 \small{
   \textbf{Correspondence:} \href{mailto:mghimire@asu.edu}{\texttt{mghimire@asu.edu}, \texttt{zhuxuan@amazon.com}}
 }
}

\newcommand\blfootnote[1]{%
  \begingroup
  \renewcommand\thefootnote{}\footnote{#1}%
  \addtocounter{footnote}{-1}%
  \endgroup
}

\begin{document}

\maketitle

\begin{abstract}
Current techniques for post-training Large Language Models (LLMs) rely either on costly human supervision or on external verifiers to boost performance on tasks such as mathematical reasoning and code generation. However, as LLMs improve their problem-solving, any further improvement will potentially require high-quality solutions to difficult problems that are not available to humans. As a result, learning from unlabeled data is becoming increasingly attractive in the research community. Existing methods extract learning signal from a model's consistency, either by majority voting or by converting the model's internal confidence into reward. Although internal consistency metric such as entropy or self-certainty require no human intervention, as we show in this work, these are unreliable signals for large-scale and long-term training. To address the unreliability, we propose PRISM, a unified training framework that uses a Process Reward Model (PRM) to guide learning alongside model's internal confidence in the absence of ground-truth labels. We show that effectively combining PRM with self-certainty can lead to both stable training and better test-time performance, and also keep the model's internal confidence in check. Code available at \href{https://github.com/ghimiremukesh/PRISM}{https://github.com/ghimiremukesh/PRISM}.
\end{abstract}
\blfootnote{$^\dagger$Work done during an internship at AWS.}
\section{Introduction}
Post-training or fine-tuning an LLM is an integral step for instilling domain knowledge into the LLM \citep{wang2025comprehensive} or aligning the LLM to desired preference \citep{ziegler2019fine}. Various techniques such as Supervised Fine-Tuning (SFT) \citep{radford2018improving}, Preference Fine-Tuning (PFT) \citep{ziegler2019fine}, Reinforcement Learning from Human Feedback (RLHF) \citep{ouyang2022training}, and Reinforcement Learning from Verifiable Rewards (RLVR) \citep{lambert2024tulu}, can be applied for post-training depending on the task. Among them, RLHF and RLVR are commonly used to post-train LLMs on mathematical and code reasoning tasks \citep{zhang2024policy,guo2025deepseek, wang2025reinforcement}, while SFT and PFT are generally used for natural language tasks such as summarization \citep{rafailov2023direct, choi2024model}. However, all of these approaches hinge on the supply of high-quality human-generated examples of the corresponding task and, as a result, have less room for scalability \citep{ouyang2022training, bai2022traininghelpfulharmlessassistant}. As LLMs increasingly get better at problem solving, curating solutions to difficult problems such as Olympiad level mathematics is difficult. Furthermore, such difficult problems may not always conform to binary verifiable rewards \citep{trinh2024solving}.

\begin{figure*}[!h]
\centering
\begin{subfigure}[t]{.4\textwidth}
  \centering
  \vspace{-1.6in}
  \includegraphics[width=\linewidth]{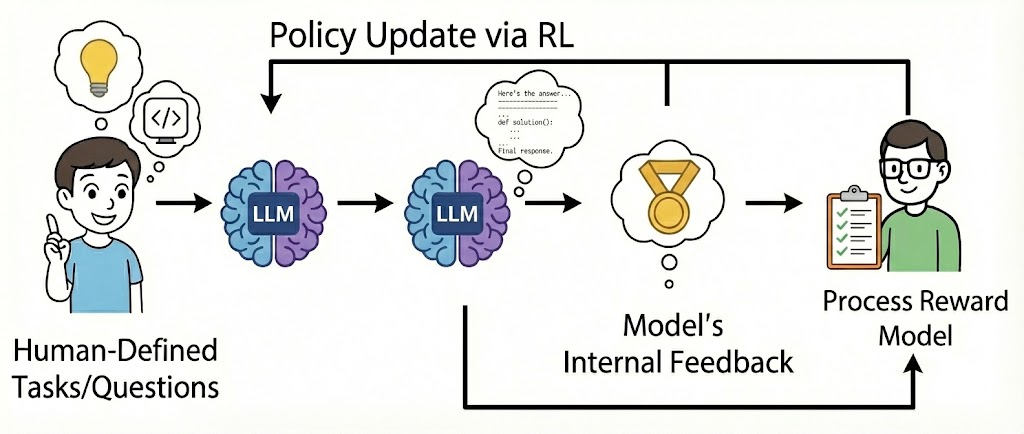}
\end{subfigure}%
\begin{subfigure}{.6\textwidth}
  \centering
  \includegraphics[width=\linewidth]{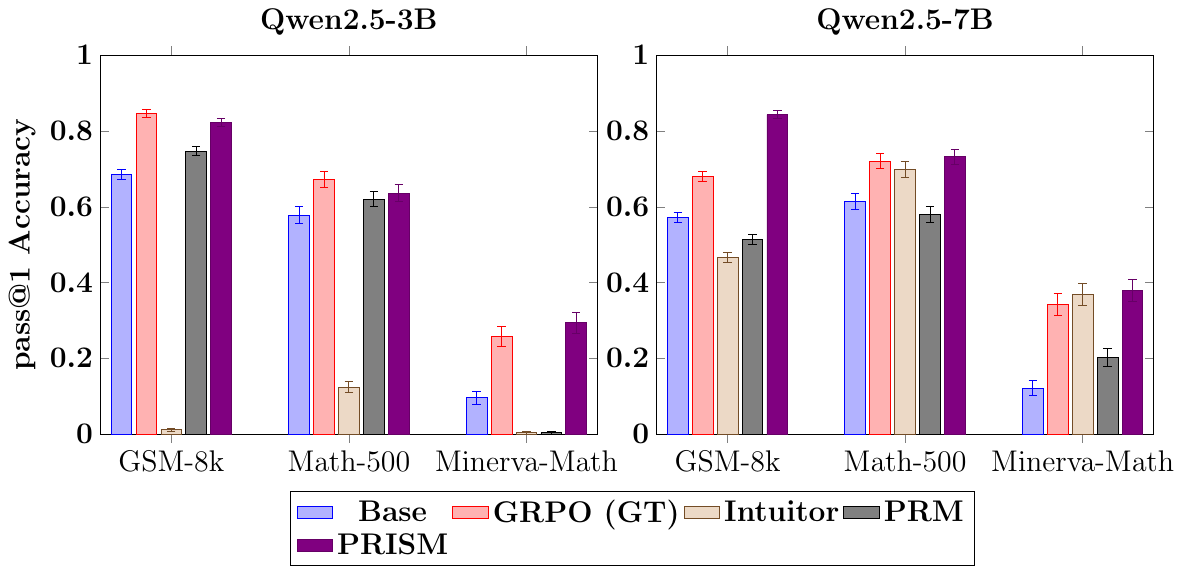}
\end{subfigure}
\caption{\textbf{Left}: An overview of our proposed PRISM framework. When ground-truth rewards are absent, LLMs learn from their intrinsic signal and utilize feedback on their reasoning process to keep its intrinsic internal signal in check. \textbf{Right:} pass@1 accuracy comparison (see Tab.~\ref{tab:acc_math} for comprehensive comparison) of Qwen2.5-3B and Qwen2.5-7B: Base, GRPO (on ground truth), INTUITOR, and PRISM (ours) on three different math benchmarks. PRISM outperforms INTUITOR, and closely matches GRPO across all benchmarks while using no ground-truth rewards; GRPO is included as a strong reference when such rewards are available.}
\label{fig:test}
\end{figure*}

Existing methods heavily rely on RLVR for post-training LLMs on mathematical and coding problems. RLVR relies on high quality domain specific verifiers, such as unit-tests and code execution environment \citep{liu2023your, liu2025code, team2025kimi, xiaomimimo} for code generation, and gold-standard solutions for math problems (tool access for assessing the correctness), and often provides a binary reward to the LLM's solution to a problem, rendering it challenging to apply to problems which lack verifiable outcomes. Recent success in solving Olympiad level math problems was attributed to general-purpose reinforcement learning and test-time compute scaling, without a heavy reliance on verifiable rewards \citep{Hu_2025}. Although the exact methodology used to achieve Gold-medal-achieving performance is not discussed, several existing works have attempted to formalize such techniques. 

To this end, \citet{self-certainty} proposes INTUITOR, that uses self-certainty reward, \citet{entropy} proposes entropy-based rewards, and \citet{zuo2025ttrl, shafayat2025can} propose the use of pseudo ground-truth labels, to enable training LLMs without any ground-truth label and rewards. While \citet{self-certainty} and \citet{entropy} use model's internal confidence as a proxy for reward, that is passed to the underlying RL algorithm, \citet{zuo2025ttrl} and \citet{shafayat2025can} utilize widely used test-time scaling (TTS) technique, i.e., majority voting, to determine a pseudo-label used to obtain binary rewards for the LLM's completions. However, a major drawback of self-consistency-based reward function is the need for human-designed answer extractor \citep{kang2025scalable}. As a result, learning algorithms that instead utilize model's internal confidence is increasingly attractive as they do not require any human intervention whatsoever \citep{self-certainty}. It is, however, unknown whether such a learning objective is sustainable for extended training, or when training on problems that the base LLM itself may not possess the adequate knowledge. 

In this work, we investigate the reliability of model's internal confidence as learning signal. Particularly we study how likely it is that the RL algorithm learns to reward-hack when the learning signal is purely based on internal-confidence. Our research question can be summarized as below:
\begin{quote}
    \textit{Can we rely solely on LLMs self-generated intrinsic reward as a learning signal to enhance their reasoning abilities on difficult math and coding problems?}
\end{quote}

We conduct a series of experiments using different \textit{Reinforcement Learning from Internal Feedback} (RLIF) frameworks, namely, self-certainty and entropy-based methods, and argue that such internal confidence based approaches artificially inflate the model's confidence irrespective of the correctness of its generation. We further show that despite the unreliableness, proxies such as self-certainty, can still be a useful learning signal when combined with an external reward. Our major contributions from this work can be summarized as below:
\begin{itemize}[noitemsep]
    \item  We show that widely used RLIF heuristics, such as self-consistency, token-entropy, and confidence-based rewards, exhibit only weak correlation with ground-truth accuracy (see Fig.~\ref{fig:rlif_corrs}).
    \item  We propose PRISM, a Process Reward \& Internal Signal Mechanism that leverages both a process-level reward model and the model’s own internal feedback to deliver stable, label-free training.
    \item When trained on various datasets (Math~\citep{MATH}, and DAPO-17k~\citep{dapo-17k}), PRISM eliminates the collapse modes seen in INTUITOR and other RLIF baselines, boosting average performance across three different evaluation benchmarks: MATH-500~\citep{math-500}, GSM-8k~\citep{gsm8k}, and Minerva-Math\footnote{https://huggingface.co/datasets/math-ai/minervamath} by approximately 34\% over the INTUITOR.
\end{itemize}

The rest of the paper is structured as follows: preliminary definitions are presented in Sec.~\ref{sec:prelim}. Experiments highlighting the failure of RLIF methods and our proposed solution are discussed in Sec.~\ref{sec:methods}. Results on various benchmarks using PRISM and other baselines are provided and discussed in Sec.~\ref{sec:results}. Finally, discussion about the limitations and future works are done in Sec.~\ref{sec:conclusion}. Related works are discussed in Appendix~\ref{sec:related_work}.
    


\section{Preliminaries} \label{sec:prelim}
In this section, we describe some of the important definitions and preliminaries. We denote $\pi_\theta(y_i|\boldsymbol{x}, \boldsymbol{y}_{<t})$ as a behavioral policy of an LLM parameterized by $\theta$, such that $\pi_\theta(\boldsymbol{y}|\boldsymbol{x}) = \prod_t \pi_\theta(y_i|\boldsymbol{x}, \boldsymbol{y}_{<t})$, where $\boldsymbol{x}$ denotes the input prompt, $\boldsymbol{y} = \{y_1, \dots y_{|\boldsymbol{y}|}\}$ denotes the response, and $\boldsymbol{y}_{<t}$ denotes prefix generated before step $t$. This behavioral policy represents a probability distribution over a vocabulary $\mathcal{V}$. 
\subsection{Group Relative Policy Optimization (GRPO)}
GRPO~\citep{shao2024deepseekmath} has recently gained popularity in the context of fine-tuning LLMs, especially due to its efficiency brought about by the removal of the value (also known as the critic) network, common in the traditional policy gradient algorithms. Instead of learning the value $v_\pi$ and the policy $\pi$  simultaneously, GRPO approximates the value via a ``group-mean'' of the the reward of $K$ samples of responses $\boldsymbol{y}_1,\dots, \boldsymbol{y}_K$ given an initial prompt $\boldsymbol{x}_0$. Given an initial behavioral policy $\pi_{\theta_0}$, and a reward function $r: \mathcal{Y} \rightarrow \mathbb{R}$, GRPO solves the following optimization problem to obtain a new policy $\pi_{\theta_1}$
\begin{equation}
\begin{aligned}
    &\max_\theta J(\theta):=\mathbb{E}_{\boldsymbol{x}_0 \sim \mathcal{X}_0,\{\boldsymbol{y}_i\}_{i=1}^k \sim \pi_{\theta_0}\left(\cdot|\boldsymbol{x}_0\right)}\\
    &\frac{1}{K}\sum_{i=1}^K \frac{1}{|\boldsymbol{y}_i|}\sum_{t}^{|\boldsymbol{y}_i|}\Bigg\{\min\left[\frac{\pi_\theta(\boldsymbol{y}_{i,t}|\boldsymbol{x}_0, \boldsymbol{y}_{i,<t})}{\pi_{\theta_0}(\boldsymbol{y}_{i,t}|\boldsymbol{x}_0, \boldsymbol{y}_{i,<t})}\hat{A}_{i, t},\right.\\
    &\left.\texttt{clip}\left(\frac{\pi_\theta(\boldsymbol{y}_{i,t}|\boldsymbol{x}_0, \boldsymbol{y}_{i,<t})}{\pi_{\theta_0}(\boldsymbol{y}_{i,t}|\boldsymbol{x}_0, \boldsymbol{y}_{i,<t})}, 1-\varepsilon, 1 + \varepsilon\right)\hat{A}_{i, t}\right] \\
    &\hspace{1.25in}-\beta\;\mathbb{D}_{\text{KL}}(\pi_\theta || \pi_{\text{ref}})\Bigg\},
\end{aligned}
\end{equation}
where $\hat{A}_{i, t} = \frac{r_i - \texttt{mean}(r_1, \dots, r_K)}{\texttt{std}(r_1, \dots, r_K)}$, is the advantage credited to all tokens in the response $y_i$. $\varepsilon$ is the threshold for the clipping, and $\beta$ is the weight for the KL penalty computed against a reference model's policy $\pi_{\text{ref}}$, which is often the base model.  

\subsection{Internal Confidence Metrics}
As discussed earlier, in this work, we primarily focus on three different internal confidence metrics which are used in the RLIF framework. 

\textbf{Token-Entropy} is the entropy of the policy conditioned on previous tokens at each step. When used as a reward function, a scalar reward is obtained by summing \citep{entropy} or taking a mean \citep{nofreelunch} over all the per-token entropy and then averaged over all responses. In this work, we use the following definition of token-entropy reward. 
\begin{equation}
    r_{\text{token-entropy}}(\boldsymbol{y}\mid \boldsymbol{x}) := -\frac{1}{|\boldsymbol{y}|}\sum_{t=1}^{|\boldsymbol{y}|}\mathcal{H}(\pi_\theta(\cdot | \boldsymbol{x}, \boldsymbol{y}<t)),
\end{equation}
where $\mathcal{H}\big(\pi_\theta(\cdot \mid \mathbf{x}, \mathbf{y}_{< t})\big)
:= - \sum_{v \in \mathcal{V}} \pi_\theta(v \mid \mathbf{x}, \mathbf{y}_{< t}) 
\log \pi_\theta(v \mid \mathbf{x}, \mathbf{y}_{< t})$ is the Shannon entropy. Since the goal is to minimize the token-level entropy, the reward is weighted negatively. 

\textbf{Trajectory-Entropy}, on the other hand, is simply the total log-probability of the LLM's response. The following trajectoy-entropy reward is used in this work, which is borrowed from \citet{nofreelunch}, which in turn is modified from \citet{entropy}'s definition. 
\begin{equation}
    r_{\text{traj-entropy}}(\boldsymbol{y}|\boldsymbol{x}) := \frac{1}{|\boldsymbol{y}|}\log\pi_\theta(\boldsymbol{y}|\boldsymbol{x})
\end{equation}
Despite the similarity in the trajectory- and token-entropy, their effect on the model's learned policy is different. Minimizing trajectory entropy results in responses with lower entropy over the trajectories, whereas minimizing token-entropy results in model generating low entropy text at each step of the generation. 

\textbf{Self-Certainty} is defined as the average KL divergence between a target distribution and the model's behavioral policy. While target distribution can be any distribution, \citet{kang2025scalable} propose to use uniform distribution $U$, which also appears in \citet{self-certainty}'s INTUITOR. A key difference between entropy and self-certainty is that the former considers a reverse KL divergence from the Uniform distribution while the latter considers the forward KL, making it mode-seeking rather than mode-covering.
\begin{equation}
    r_{\text{self-certainty}}(\boldsymbol{y}|\boldsymbol{x}) := \frac{1}{|\boldsymbol{y}|}\sum_{t=1}^{|\boldsymbol{y}|}\mathbb{D}_{\text{KL}}(U||\pi_\theta(\cdot|\boldsymbol{x}, \boldsymbol{y}_{<t}))
\end{equation}
\section{Methods} \label{sec:methods}
In this section, we first highlight the limitations of RLIF and provide the motivation for developing a better approach to learning without ground-truth labels. We then introduce our proposed method, PRISM, which enables stable training while providing a good approximation of true accuracy in the absence of ground-truth verification.

\subsection{RLIF Fails Under Extended Training} \label{sec:failure}
As shown in \citet{liu2025prorl}, reasoning space of an LLM can be sufficiently expanded through extended RL training steps. Motivated by this idea, we first put to test the existing RLIF algorithms under prolonged training. While significant performance improvement was achieved by \citet{self-certainty} training Qwen models on MATH dataset just for 1 epoch, no further analysis on extended training is provided. Keeping the hyperparameters same as in \citet{self-certainty}, we train three RLIF methods for 300 optimization steps ($\approx$6 epochs), with each method consisting of one of the three internal rewards: token entropy, trajectory entropy, and self-certainty. We monitor the true mean accuracy (comparing with the ground-truth solution) of the training rollouts as well alongside their mean lengths.

\begin{figure}[!h]
    \centering
    \includegraphics[width=\linewidth]{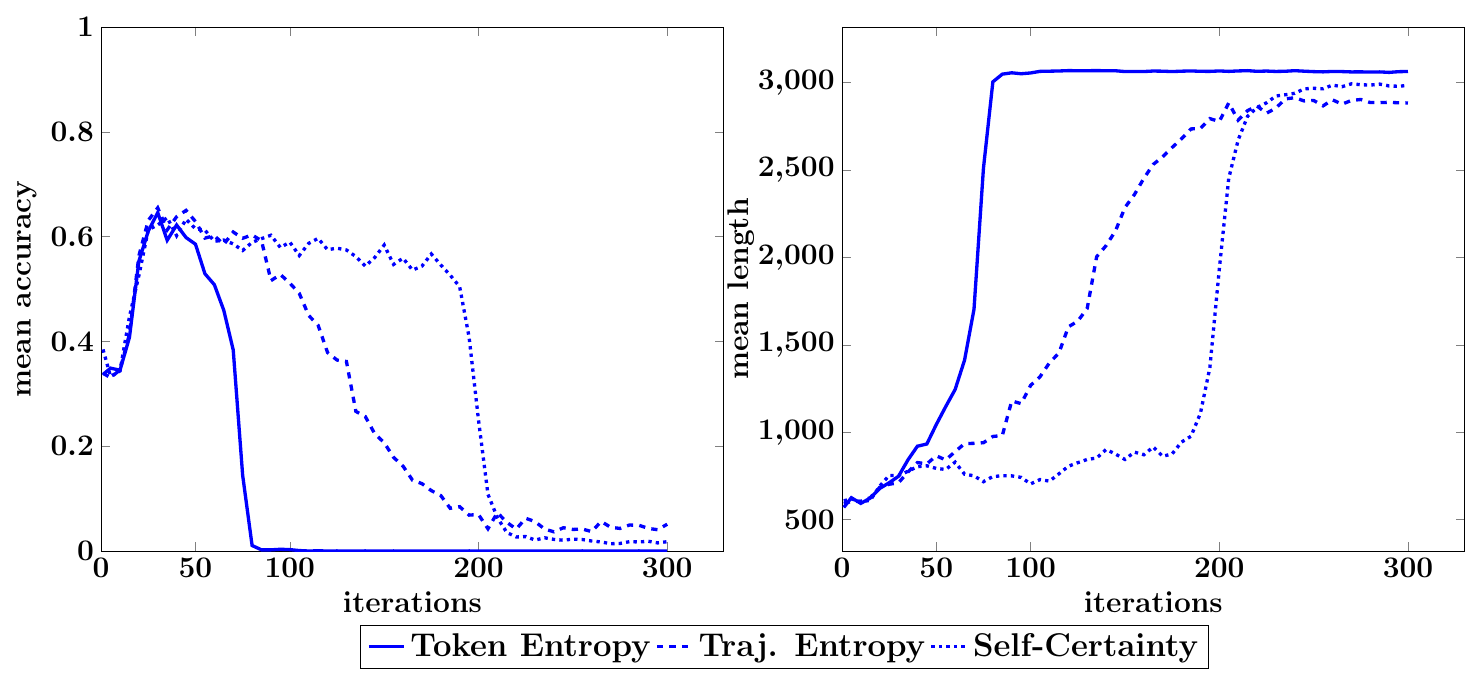}
    \caption{Mean accuracy (left) and mean length (right) of the training rollouts under three different RLIF methods. Initial trend shows rapid increase in mean accuracy across all methods, but the models start to degrade as the training progresses. Mean length of the generations start to increase and correspond to the step when the accuracy starts to drop. The base model is Qwen2.5-3B and the training data is MATH.}
    \label{fig:rlif_fail}
    \vspace{-0.2in}
\end{figure}

Results in Fig.~\ref{fig:rlif_fail} show that all three methods result in a drastic improvement in mean accuracy in the first epoch of the training. However, as the training progresses, token-entropy based RLIF fails the earliest, followed by trajectory entropy and finally self-certainty. Correlated to the mean accuracy is the mean length of the completion. This correlation suggests that as the training progresses, the model learns to inflate its reward by generating confident responses leading to higher reward -- a classic sign of reward hacking. Upon close inspection (see Appendix~\ref{app:fail}) we observe that while self-certainty-trained model learns to append new unrelated questions to its generation to demonstrate its confidence, the entropy-based methods simply repeat the generation without solving the problem. Among the three, entropy-based methods seem to severely impact the model's reasoning performance. Since self-certainty based method demonstrates less severe failure mode than entropy based methods, we use INTUITOR as a baseline for comparison against our proposed method in the later sections. 
\vspace{-0.1in}
\subsection{Internal Proxy Rewards are Unreliable Signals}
The RLIF frameworks that we introduced use entropy and self-certainty as proxy rewards. In a case where there is truly no ground-truth solution to evaluate the model's performance during training, it becomes difficult to assess the training performance on the target problem distribution. Sec.~\ref{sec:failure} showed degrading performance, which is only the half of the story. To better understand how the learning signal (i.e., the proxy rewards) track with the true accuracy, we monitor the mean rewards for the three RLIF approaches.

\begin{figure}[!h]
\begin{subfigure}[t]{\linewidth}
  \centering
  \includegraphics[width=\linewidth]{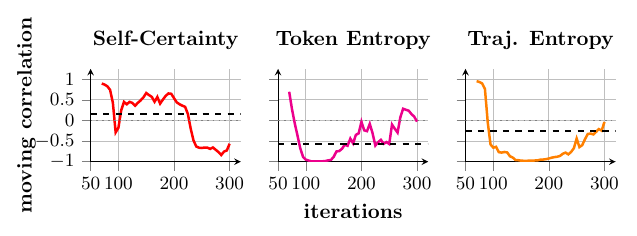}
\end{subfigure}
\begin{subfigure}{\linewidth}
  \centering
  \includegraphics[width=\linewidth]{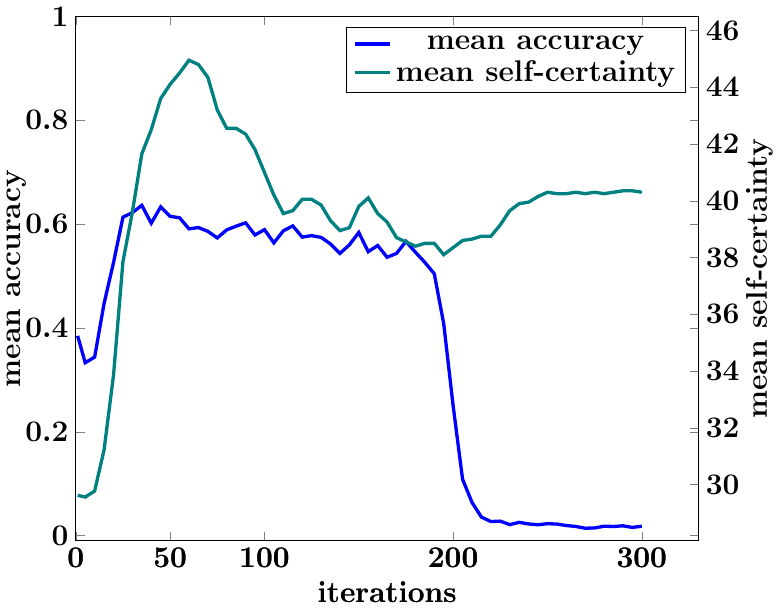}
\end{subfigure}
\caption{\textbf{Top:} Rolling correlation between true accuracy and the respective proxy rewards during training across all methods. All training were run for total of 300 optimization steps ($\approx$ 6 epochs) on MATH dataset. Dashed lines denote mean correlation. \textbf{Bottom:} Mean accuracy vs mean self-certainty of training rollouts. Self-certainty score does not collapse with the true accuracy.}
\label{fig:rlif_corrs}
\end{figure}

We plot the moving correlation between the mean internal-feedback rewards (self-certainty, token, and trajectory entropy) and true accuracy in Fig.~\ref{fig:rlif_corrs}. Results show almost no correlation between these two signals, which is consistent to our findings in Sec.~\ref{sec:failure}. The implication of these results is that it becomes fundamentally difficult to monitor training stability in true label-free learning. 
\subsection{Process Reward Models as an Alternative}
PRMs such as GenPRM~\citep{zhao2025genprm} can be a good alternative or addition to the existing RLIF framework for potentially addressing the limitations highlighted above. We briefly introduce GenPRM and explain how it fits into the RLIF framework. For clarity, we will denote the base model that we train as a Student model.

\textbf{GenPRM} is a generative process reward model proposed by \citet{zhao2025genprm}, which unlike previous methods, performs explicit chain-of-thought (CoT) reasoning, with code verification capabilities, to provide comprehensive feedback. It is \textit{generative} and not \textit{discriminative} due to its ability to perform CoT before providing a judgement rather than a direct binary 0/1 feedback commonly provided by discriminative models. 

In the standard TTS implementation of GenPRM, the response $\boldsymbol{y}$ of the Student model $\pi_\theta$ to a question $\boldsymbol{x}$ is first split into individual answer paragraph (step) $a_1, \dots, a_M$ and sequentially fed into the GenPRM model $\pi_\psi$ to produce a reasoning $v_m$, and a judgement $j_m$ in the form of \textbf{Yes} or \textbf{No} token: 
\begin{align*}
\small
v_m &= \pi_\psi\Big(\cdot|\boldsymbol{x}, \{a_1, v_1, j_1, \dots, a_{m-1}, v_{m-1}, j_{m-1}\},\\
&\hspace{2.5in} a_m\Big)\\
j_m &= \pi_\psi\Big(\cdot|\boldsymbol{x}, \{a_1, v_1, j_1, \dots, a_{m-1}, v_{m-1}, j_{m-1}\},\\
&\hspace{2.5in} a_m, v_m\Big)
\end{align*}
Finally, the reward $r_m$ to the step $m$ is:
\begin{equation}\label{eq:prm_reward}
\begin{aligned}
    r_m &= \pi_\psi\Big(\textbf{Yes}|\boldsymbol{x}, \{a_1, v_1, j_1, \dots,\\
    &\hspace{0.8in}a_{m-1}, v_{m-1}, j_{m-1}\}, a_m, v_m\Big).
\end{aligned}
\end{equation}
The reward computation can be further refined by averaging
\eqref{eq:prm_reward} over $N$ separate GenPRM generations (or calls). 

Once all paragraphs are graded, the final reward to the Student model's generation is the aggregation of all step-wise rewards $\{r_m\}_{m=1}^M$. 
\begin{equation}
r_G = \texttt{agg}_m \{r_m\}_{m=1}^M, 
\end{equation}
where $\texttt{agg}$ is the aggregation function of choice, such as $\texttt{min}$, $\texttt{max}$, $\texttt{mean}$. We use the $\texttt{min}$ aggregator in this work, as we found it to be more stable and effective than $\texttt{mean}$ in our preliminary tests.

In practice, instead of manaully splitting $\boldsymbol{y}$ into chunks of paragraphs $a_m$, we directly provide the entire raw response $\boldsymbol{y}$ from the Student model into GenPRM and prompt it to provide judgements $j_m$ for each paragraph the GenPRM recognizes (see App.~\ref{app:genprm_details} for details). This is necessary due to the heavy computational overhead caused by sequential reward assignment, which also makes it difficult to parallelize. As a result, we also prompt GenPRM to provide a completion judgement $c$ with a single \textbf{Yes} or \textbf{No} verdict in the end computed similarly as in \eqref{eq:prm_reward}. The final reward to the generation is then a harmonic mean of aggregated process reward and completion reward. 
\begin{equation}
    r_G = \frac{2 \times \texttt{agg}_m \{r_m\}_{m=1}^M \times r_c}{\texttt{agg}_m \{r_m\}_{m=1}^M + r_c}
\end{equation}
Moving forward, we use a short-hand $\text{PRM}(\boldsymbol{y}|\boldsymbol{x})$ to denote the reward received from the PRM model. With that, we are ready to present our preliminary results on using GenPRM as a reward signal instead of internal feedback. To highlight the effectiveness and reliableness of PRMs, we first test the verification ability of each of the methods, i.e., how well do each of the rewards recognize correct vs incorrect answers. This can be studied by plotting the distribution of rewards assigned to correct and incorrect generations. For the remainder of the study, we only use self-certainty (INTUITOR) as the RLIF framework for comparison as it demonstrated the best performance out of the three RILF proxy rewards. 

\begin{figure*}[!h]
    \centering
    \includegraphics[width=\linewidth]{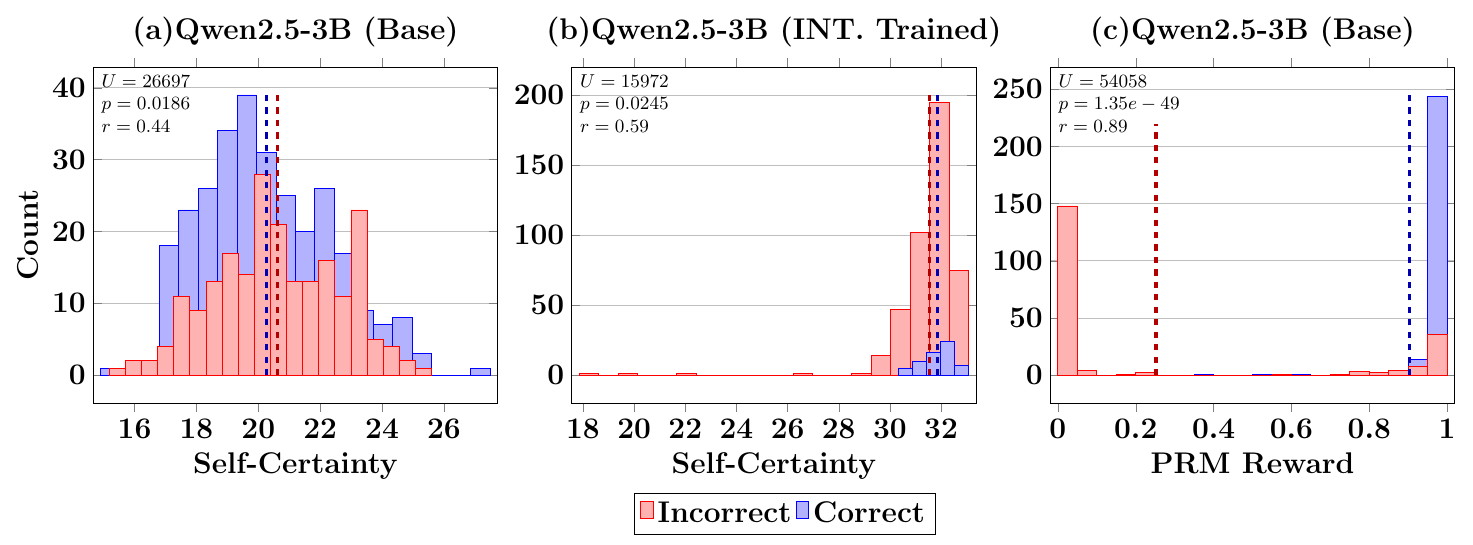}
    \caption{Distribution of self-certainty scores of responses generated by \textbf{(a)} Qwen2.5-3B base model and \textbf{(b)} Qwen2.5-3B INTUITOR trained model. \textbf{(c)} Distribution of PRM rewards for responses generated by Qwen2.5-3B base model. $U$ is the Mann-Whitney U-test score which quantifies the separation between two distributions. $p$ and $r$ are the $p$-value and effect-size respectively. Unlike self-certainty, PRM rewards reliably predict correct from incorrect responses. Dashed line represent mean of the respective distribution.}
    \label{fig:score_dist_first}
\end{figure*}

Fig.~\ref{fig:score_dist_first} shows the distribution of respective rewards for correct and incorrect responses to the problems in Math-500 dataset. Qualitatively, it is apparent that we cannot rely on self-certainty scores to distinguish between correct and incorrect responses, which is quantitatively supported by Mann-Whitney $U$ scores~\citep{shier2004statistics} for each of the score distributions along with their effect sizes and $p$-values. PRM, as shown in Fig.~\ref{fig:score_dist_first}(c), more often than not scores correct and incorrect responses appropriately. Furthermore, from Fig.~\ref{fig:score_dist_first}(b), we observe that INTUITOR, the RLIF method that uses self-certainty reward, increases overall confidence of the model regardless of whether the response is correct or not, which is its yet another limitation: \textit{RLIF promotes over-confidence.}

\begin{figure}[!h]
    \centering
    \includegraphics[width=\linewidth]{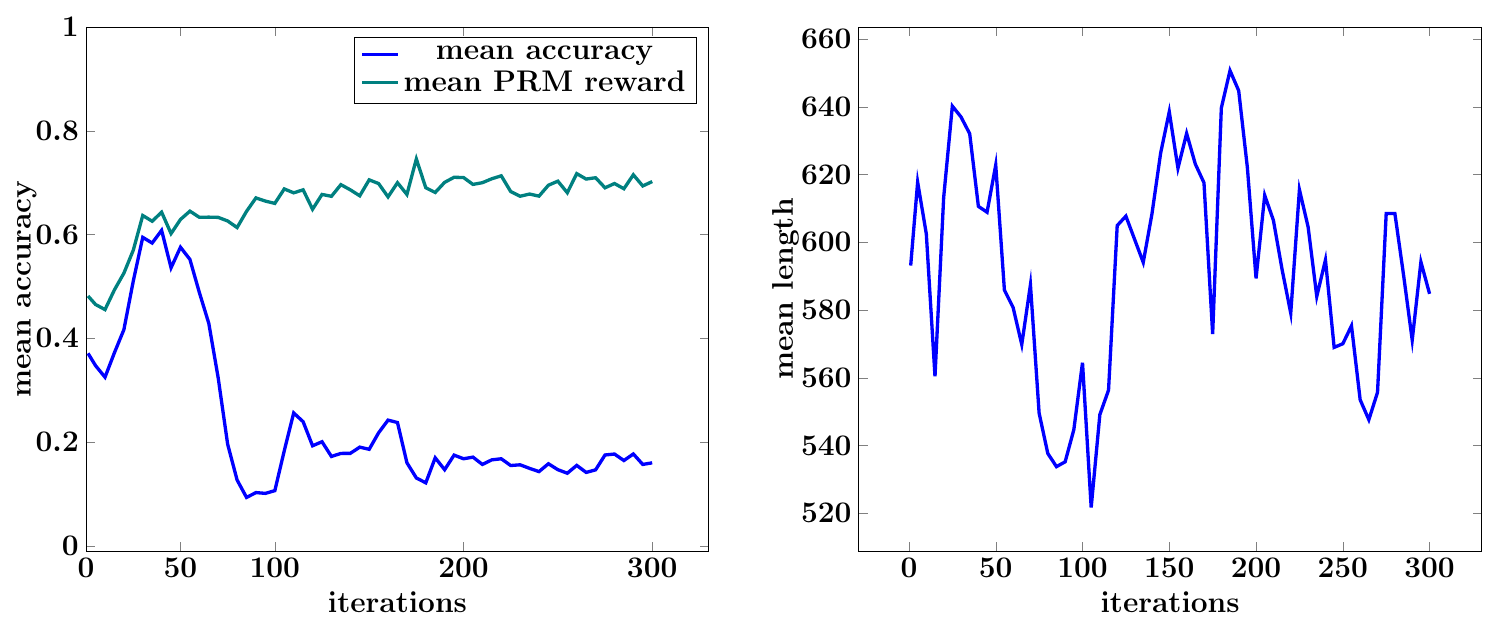}
    \vspace{-0.25in}
    \caption{\small (left) Mean accuracy and PRM rewards of the training rollouts and (right) mean length of rollouts.}
    \label{fig:prm_rollout}
\end{figure}

Due to its demonstrated verification accuracy, we propose to replace the self-certainty reward in the RLIF framework with PRM reward. We repeat the experiment in Sec.~\ref{sec:failure} for 300 optimization steps. Similar to Fig.~\ref{fig:rlif_fail}, we plot the mean accuracy and mean PRM rewards for the training rollouts in Fig.~\ref{fig:prm_rollout}, as well as the mean length of the rollouts. We observe that similar to RLIF rewards, the accuracy initially increases followed by sudden drop in accuracy, but increasing PRM rewards. However, a surprising observation is that the mean length remains stable. Upon close inspection, we notice that as the training progresses, the model ``forgets'' to box (\textbackslash$\texttt{boxed}$\{\}) its final solution, despite the instruction (system prompt) specifying it to do so. Since the ground-truth verifier expects the final solution to be boxed, the accuracy reward for generations not containing a box is zero. An example of such rollout during training is shown in Fig.~\ref{fig:nobox}. 

While adding an additional penalty for format adherence is an option, it would require task-specific format, demanding human supervision. Unlike, PRM-reward trained model, INTUITOR trained model quickly demonstrated instruction following tendencies. We conduct additional experiments (see App. \ref{sec: effect}) to better understand the benefits of PRM and self-certainty rewards. The results show that self-certainty reward promotes instruction-following better than PRM rewards, and PRM rewards on the other hand, prevents the model from begin over-confident. 




\subsection{A Unifying Solution}
Because of self-certainty's ability to quickly push the model to learn the response structure, we propose to complement PRM rewards with self-certainty rewards. 
For an input prompt $\boldsymbol{x}$, let $\boldsymbol{y}$ be the model's generation. The response $\boldsymbol{y}$ is then scored as:
\begin{equation}
    r^{sc} = r_\text{self-certainty}(\boldsymbol{y} | \boldsymbol{x}), \quad r^{PRM} = \text{PRM}(\boldsymbol{y} | \boldsymbol{x})
\end{equation}
The advantages are then computed for each of the rewards separately based on the group mean rewards. 

\begin{equation}
\begin{aligned}
    &\hat{A}_{i, t}^{sc} = \frac{r^{sc}_i - \texttt{mean}(\{r^{sc}_1, \dots, r^{sc}_K\})}{\texttt{std}(\{r^{sc}_1, \dots, r^{sc}_K\})},\\
    &\hat{A}_{i, t}^{PRM} = \frac{r^{PRM}_i - \texttt{mean}(\{r^{PRM}_1, \dots, r^{PRM}_K\})}{\texttt{std}(\{r^{PRM}_1, \dots, r^{PRM}_K\})},
\end{aligned}
\end{equation}
The final advantage is then computed as follows:
\begin{equation}
    \hat{A}_{i, t} = \gamma \hat{A}_{i, t}^{sc} + \hat{A}_{i, t}^{PRM},
\end{equation}
where $\gamma$ is the weight to the self-certainty advantage, and $\gamma=0$ corresponds to the case with just PRM advantages as learning signal. Because this leverages both process reward and model's internal confidence signal, namely the self-certainty, we call it a \textbf{P}rocess \textbf{R}eward \& \textbf{I}nternal \textbf{S}ignal \textbf{M}echanism (PRISM).

\section{Experiments and Results} \label{sec:results}
\subsection{Experimental Setup}
We use Open-R1 framework \citep{openr1}, and GRPO as the RL algorithm across all the methods: GRPO with Ground-Truth, INTUITOR, and PRISM. As base models, we use Qwen2.5-3B and Qwen2.5-7B \citep{qwen}, that are pre-trained to mimic a chat style conversation. We use GenPRM-7B~\citep{zhao2025genprm} as the process reward model in PRISM. For math reasoning problems, Qwen2.5-3B was trained on MATH dataset \citet{MATH} containing 7500 problems, and Qwen2.5-7B was trained on the \texttt{en} subset of the DAPO-17k \cite{dapo-17k} dataset containing difficult math problems.
Different hyperparameters such as the gradient accurumulation steps, per device batch size, and number of generations, that determine the overall training batch size per update step are tabulated in App.~\ref{app:hyper}. We select the hyperparameters so that the overall batch size and the total unique prompts at each step are roughly equal across all methods. On the math datasets, we generate 6-7 generations per problem with Qwen2.5-3B, and 16 generations per problem with Qwen2.5-7B. Following \citet{self-certainty}, KL penalty of $\beta=0.005$ is used when training on math problems. Following INTUITOR's implementation, a cosine schedule is used to decay the learning rate. All trainings were performed on NVIDIA A100 GPUs with 40 or 80 GB (only for Qwen2.5-3B PRISM and PRM) memory. 

\subsection{Evaluation}
We use \texttt{lighteval} \citep{lighteval} as the standard evaluation suite for evaluating accuracy of the models on math reasoning task. We use two popular benchmark datasets: Math-500 \citep{math-500}, GSM-8k \citep{gsm8k}, and Minerva-Math. We use the pass@1 (1 sample) score to report the accuracy of the generations. All generations during evaluation use greedy decoding. To evaluate on code reasoning, we use LiveCodeBench (LCB) \citep{lcb} as the evaluation suite, and evaluate the accuracy on python code generation. The evaluations were performed on NVIDIA A100 GPUs with 40 GB memory. 

\begin{table*}[!h]
  \centering
  \caption{Accuracy on Math Benchmarks. All trainings were terminated after 6 epochs, and the final checkpoints were used.}
  \label{tab:acc_math}

\resizebox{0.8\textwidth}{!}{%
\begin{tabular}{@{}cccccc@{}}
\toprule
Model       & Data                                         &   Method    & GSM-8k                                       & Math-500                                     & Minerva-Math                                      \\ \midrule
\multicolumn{1}{c}{\multirow{5}{*}{Qwen2.5-3B}} & & Base  & 0.6861 ± 0.0128                              & 0.578 ± 0.0221                               & 0.096 ± 0.0179                    \\
\multicolumn{1}{c}{}                            &  & GRPO (GT)  & \textbf{0.8461 ± 0.0099}                     & \textbf{0.6720 ± 0.0210}                     & 0.257 ± 0.0265                      \\
\multicolumn{1}{c}{}             &     MATH          & INT.   & 0.0114 ± 0.0029                              & 0.124± 0.015                                 & 0.004 ± 0.0037                                \\
\multicolumn{1}{c}{}             &               & PRM   & 0.7460 ± 0.0119                              & 0.62 ± 0.02                                  & 0.004 ± 0.0037                                \\
\multicolumn{1}{c}{}             &               & PRISM & 0.8234 ± 0.0105                    & 0.636 ± 0.022                      & \textbf{0.294 ± 0.0277}                     \\ \midrule
\multirow{5}{*}{Qwen2.5-7B}      &               & Base  & \multicolumn{1}{c}{0.5709 ± 0.0136}          & \multicolumn{1}{c}{0.614 ± 0.022}            & \multicolumn{1}{c}{0.121 ± 0.0198}                \\
                                 &               & GRPO (GT)  & \multicolumn{1}{c}{0.679 ± 0.013}            & \multicolumn{1}{c}{0.72 ± 0.02}              & \multicolumn{1}{c}{0.342 ± 0.0288} \\
                                  &    DAPO-17k          & INT.   & \multicolumn{1}{c}{0.4655 ± 0.0137}          & \multicolumn{1}{c}{0.6980 ± 0.0244}          & \multicolumn{1}{c}{0.368 ± 0.0293} \\
                                 &               & PRM   & \multicolumn{1}{c}{0.5140 ± 0.0138}          & \multicolumn{1}{c}{0.58 ± 0.0221}            & \multicolumn{1}{c}{0.202 ± 0.0244}          \\
                                 &               & PRISM & \multicolumn{1}{c}{\textbf{0.8431 ± 0.0100}} & \multicolumn{1}{c}{\textbf{0.7320 ± 0.0198}} & \multicolumn{1}{c}{\textbf{0.379 ± 0.0295}}\\\bottomrule
\end{tabular}
}
\end{table*}

\subsection{Results}
\begin{figure}[!h]
    \centering
    \includegraphics[width=\linewidth]{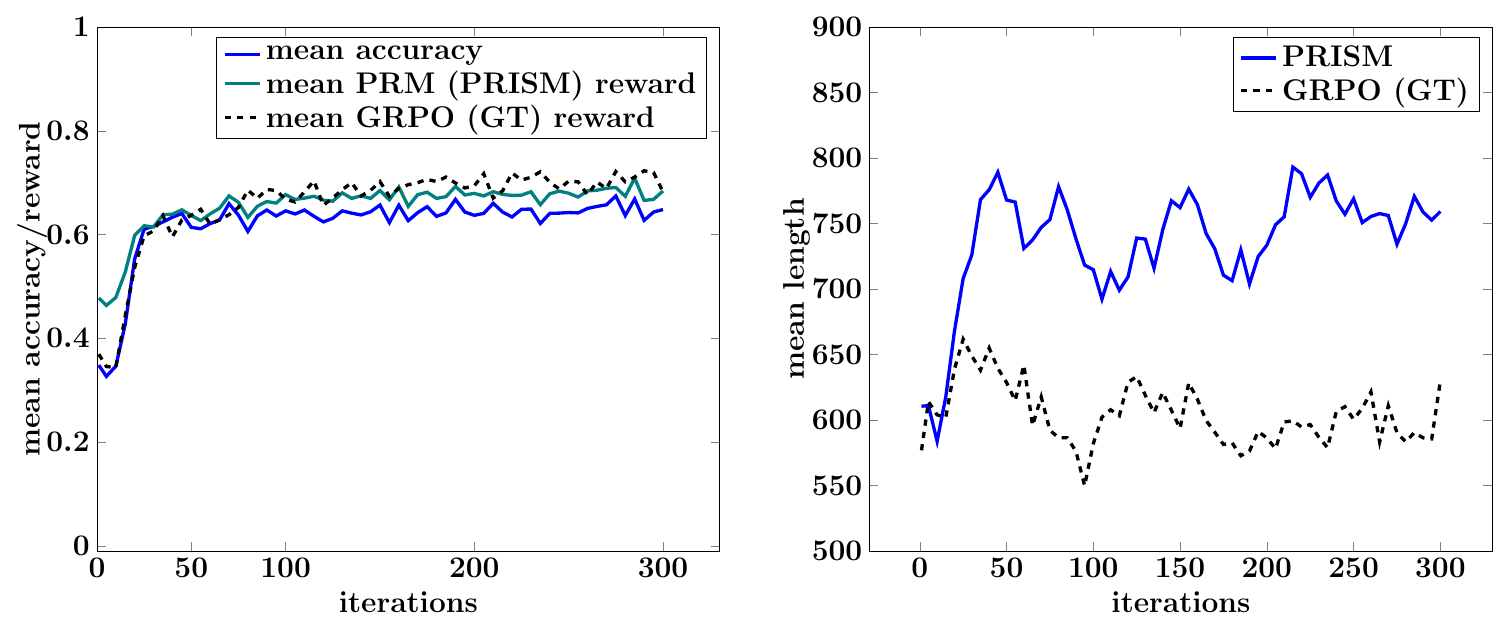}
    \caption{\textbf{Left:} Mean rewards and accuracy of the training rollouts with PRISM, and GRPO (GT). PRISM not only is stable, but also represents the actual accuracy despite lacking the ground-truth and compares similarly to the case where ground-truth labels are used. \textbf{Right:} Mean length of the rollouts. PRISM's rollouts are considerably concise and do not generate verbose response as in the case of pure RLIF (cf Fig.~\ref{fig:rlif_fail}).}
    \label{fig:prism_curve}
\end{figure}

First, we show the training performance of PRISM on Qwen2.5-3B model for math reasoning task. We use the same MATH dataset for training as in previous sections. Fig.~\ref{fig:prism_curve} visualizes the mean true accuracy and the mean process rewards from the PRM model (GenPRM-7B), as well as the mean length of the rollouts. We also report the training performance when trained with the ground-truth reward using GRPO. Note that all methods use GRPO as underlying RL algorithm. We simply use GRPO (GT) to denote the case that uses ground-truth labels for convenience. 

\begin{figure}[!h]
    \centering
    \includegraphics[width=\linewidth]{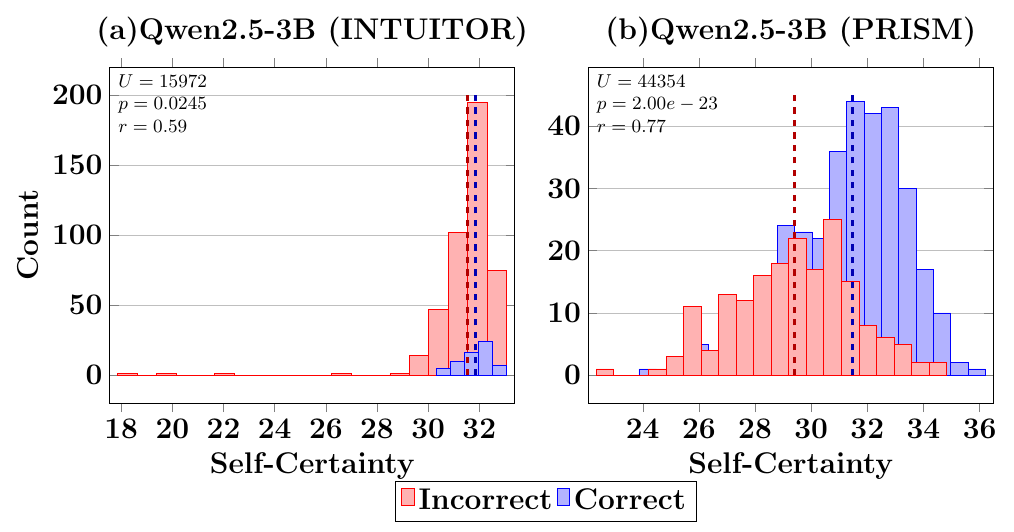}
    \caption{(a)Histogram of self-certainty scores of correct and incorrect responses to MATH-500 generated with INTUITOR trained model. (b)Histogram of self-certainty scores of correct and incorrect responses to MATH-500 generated with PRISM trained model. INTUITOR tends to increase the model's internal confidence regardless of its correctness, while PRISM helps to keep the model's confidence in check. Mann-Whitney U scores between self-certainty of correct and incorrect responses are higher for PRISM compared to INTUITOR showing better separation.}
    \label{fig:conf_check}
\end{figure}

Furthermore, we observe that PRISM is able to keep the model's internal confidence in check, unlike INTUITOR that leads to overall increase in model's confidence. Fig.~\ref{fig:conf_check} shows the distribution of self-certainty scores of the INTUITOR and PRISM trained models' generation to Math-500 problems. Self-certainty scores of PRISM trained model's generation shows better separation between its correct and incorrect response compared to that of the INTUITOR trained model's, substantiated by the Mann-Whitney $U$ scores and the mean self-certainty scores of correct and incorrect generations.
\begin{table}[!h]
\centering
\caption{Out-of-Domain Accuracy on Coding Task.}
\label{tab:out-domain}

\resizebox{0.4\textwidth}{!}{%
\begin{tabular}{@{}lccc@{}}
\toprule
\multicolumn{1}{c}{} Model  & Data     &    Method      & LCB         \\ \midrule
\multirow{5}{*}{Qwen2.5-3B} & \multirow{5}{*}{MATH} & Base     & 0.135          \\
                            & & GRPO (GT)    & \textbf{0.145}  \\
                            & & INT. & 0.129           \\
                            & & PRM & \textbf{0.145} \\
                            & & PRISM    & 0.140  \\ \midrule
\multirow{5}{*}{Qwen2.5-7B} &\multirow{5}{*}{DAPO-17k} & Base     & 0.212           \\
                            & & GRPO (GT)    & 0.2117          \\
                            & & INT. & 0.2167          \\
                            & & PRM & 0.2116 \\
                            & & PRISM    & \textbf{0.2169}\\ \bottomrule
\end{tabular}
}
\end{table}
In Table~\ref{tab:acc_math}, we present the results on in-domain task -- math reasoning for a smaller Qwen2.5-3B model and a larger Qwen2.5-7B model. Both models trained with PRISM outperform INTUITOR. While Qwen2.5-3B's performance is very close to GRPO (w/ ground-truth), Qwen2.5-7B either outperforms or ties with it across all benchmarks. Qwen2.5-7B was trained on a difficult DAPO dataset, and might have benefited from PRISM's process rewards as opposed to GRPO's sole outcome-based rewards. 

Additionally, to test the out-of-domain performance, we evaluate the code generation performance of the model trained on math reasoning task. Specifically, we test the out-of-domain performance on LCB's code generation task. The results are tabulated in Table~\ref{tab:out-domain}. PRISM outperformed INTUITOR across both model sizes, however, PRM trained (PRISM with $\gamma=0$) Qwen2.5-3B model outperformed and was comparable to GRPO (with ground-truth reward) trained model. This suggests that careful tuning of $\gamma$ might be necessary. Regardless, PRISM consistently outperformed INTUITOR. 

\section{Conclusion} \label{sec:conclusion}
In this work we introduce PRISM as a more robust alternative to RLIF, which suffers from fundamental challenges -- most notably, the unreliability of internal confidence as a reward signal and the tendecy to produce over-confident, yet, incorrect generations. Our experiments demonstrate that PRISM not only mitigates these failure modes, but also consistently outperforms RLIF baselines by combining the complementary strengths of a process-level reward model and the model's own internal feedback. PRISM improves the accuracy on three different math benchmark tasks by an average of 34\% compared to INTUITOR. 
While PRISM addresses key weaknesses of RLIF, it still inherits limitations from its reliance on learned external models. Future work will explore adaptive weighting (with $\gamma$) between internal and process-level signals, as well as online update of the PRMs. 

\section{Limitations}
While PRISM enables RL training in the absence of ground-truth labels, it still optimizes proxies rather than the ground truth, leaving room for misspecification and gaming. Crucially, the effectiveness of PRISM is upper-bounded by the capability of the underlying PRM. If the PRM fails to detect reasoning errors or hallucinates critiques, the policy may converge to incorrect reasoning patterns.

Furthermore, our evaluation centers on math and code reasoning with static benchmarks; generalization to open-ended, multi-modal, or interactive settings remains untested. We also note that utilizing a generative PRM for reward calculation introduces higher inference latency during the training rollout phase compared to lightweight metrics like self-consistency or entropy. Finally, the proposed approach does not provide formal guarantees against reward hacking or long-horizon failure modes; when applied beyond our evaluation scope, proxy optimization can yield persuasive but incorrect outputs and amplify biases.

\bibliography{reference}


\appendix
\onecolumn

\section{Appendix}
\subsection{Related Work} \label{sec:related_work}

\subsubsection{Test-Time Scaling and Majority Voting}
Test-time scaling refers to the use of compute during the inference stage to generate higher quality response from the LLM. Popular test-time scaling approach include Self-Refine \citep{madaan2023self}, Best-of-N \citep{snell2024scaling}, Majority Voting \citep{wang2022self}. Self-Refine works by iteratively prompting the LLM to revise its generation depending on the compute resources or until the LLM deems its response is adequate. It has been shown that this iterative cycle of revision can elicit better performance in natural language and code generation tasks. Best-of-N, on the other hand allows an LLM to generate $N$ completions for a given prompt and selects the best response based on some predefined heuristics. Akin to this is the idea of majority voting, which also requires an LLM to generate $N$ responses and selects the most repeated response. Best-of-N and Majority Voting are often referred to as Consistency-based method, as they are a proxy to how consistent the model is in its response. 

With test-time scaling framework, the compute used to generate the ``optimal'' response is often discarded, which can instead be used as a feedback to the model. This learning paradigm is termed as Test-Time Learning (TTL), and has recently been applied as an RL framework to improve the model online during inference \citep{zuo2025ttrl}. Because ground truth data are assumed to be not available during inference, test-time learning frameworks fall under the broader framework of ``learning without ground-truth''. A primary challenge of consistency-based approach in TTL is the need to determine the convergence of a response to a single answer, which is often difficult in the case of open-ended generations. 

\subsubsection{Internal Confidence}
Another approach to learning without ground-truth reward is the paradigm of \textit{Reinforcement Learning from Internal Feedback} (RLIF). Under this framework, the model is optimized to maximize its internal confidence metric without any external supervision or ground-truth rewards. Common internal-confidence metrics include: token-level and trajectory-level entropy~\citep{entropy} and self-certainty~\citep{kang2025scalable}. In essence, all of these methods minimize entropy in the sense that the probabilities gradually shift towards one-hot, albeit their interpretations are different. Other similar approaches exist that work on the semantic space. For example, \citet{zhang2025right} proposes Entropy Minimized Policy Optimization (EMPO), which minimizes LLM's entropy on the questions in it's latent semantic space, while \citet{chen2025seed} proposes SEED-GRPO, which on top of ground truth rewards, minimizes the entropy of the model's generation. 

\subsubsection{Process Reward Models (PRMs)}
Reward models (RMs) are central to preference optimization, where user preferences are collected and used to train a model that scores the LLM's generation. The scores obtained from the reward model are then used to refine the model via RL~\citep{ouyang2022training}. RMs are limited to outcome-based scoring, focusing mainly on the end result without taking into account intermediate reasoning~\citep{math-500}. Unlike sparse binary rewards from RMs, PRMs offer a dense process rewards, which provide feedback on intermediate reasoning steps of the LLMs' response, and have been shown to be effective in test-time scaling of LLMs \citep{uesato2022solving, math-500, yuan1981free, wang2023math}. Among the available PRMs, GenPRM \citep{zhao2025genprm} has been shown to be effective for process supervision, and verification for math and code reasoning problems. Apart from its superior performance in verification tasks, \citet{zhao2025genprm} demonstrate GenPRM's ability to serve as a critique for improving test-time performance of a model via iterative refinement.

\subsection{Effect of Self-Certainty Reward} \label{sec: effect}
\begin{figure}[!h]
    \centering
    \includegraphics[width=0.75\linewidth]{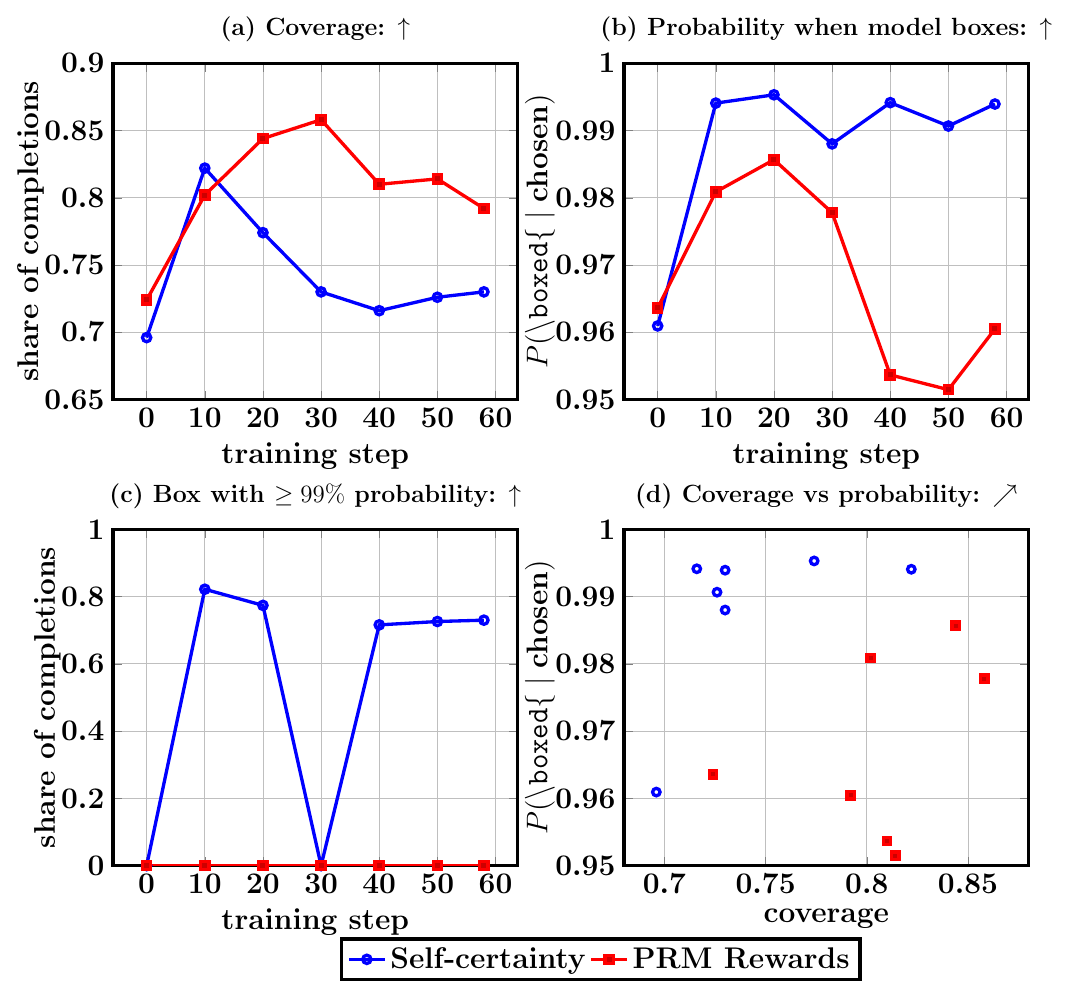}
    \caption{(a) Frequency of boxed answers on Math-500 problems. (b) Probability with which the model boxes the answer when it does so. (c) Frequency of boxes with $\ge 99\%$ probability. (d) Probability vs frequency plot with top-right being better. Both methods used Qwen2.5-3B as the base model and were trained for 1 epoch.}
    \label{fig:boxed_probs}
\end{figure}

To study this, first we plot the frequency of boxed responses and their probabilities at various checkpoints on Math-500 dataset. Fig.~\ref{fig:boxed_probs} shows that INTUITOR (RLIF with self-certainty) consistently boxes with higher probability as opposed to PRM-reward trained model.

\begin{figure}[!h]
\centering
\begin{subfigure}{\textwidth}
    \includegraphics[width=\linewidth]{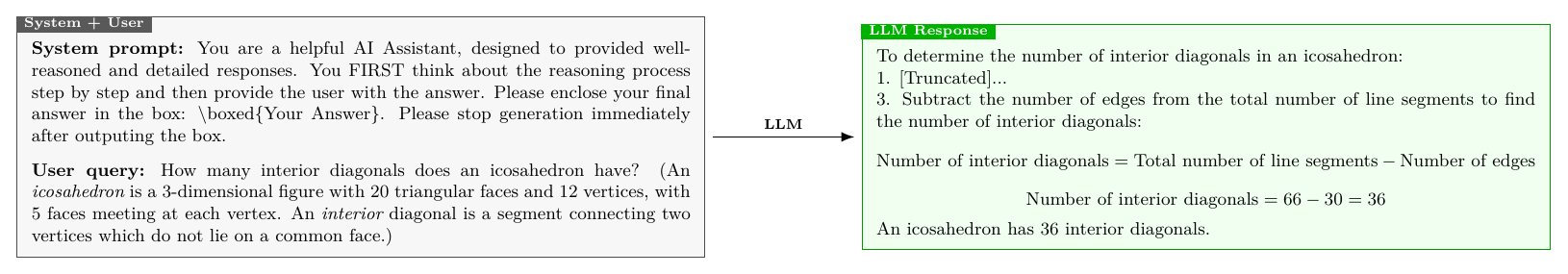}
    \caption{Qwen2.5-3B model trained with PRM rewards forgets to box the answers causing accuracy to be zero.}
    \label{fig:nobox}
\end{subfigure}\hfill
\begin{subtable}{0.6\textwidth}
  \centering
  \caption{Freq. of transition words (MATH-500)}
  \label{tab:frequency}
  \begin{adjustbox}{max width=\linewidth}
  \begin{tabular}{@{}lccccccc@{}}
  \toprule
                 & step-0 & step-10 & step-20 & step-30 & step-40 & step-50 & step-58 \\ \midrule
  Self-Certainty & 0.092  & 0.126   & 0.11    & 0.098   & 0.074   & 0.076   & 0.076   \\
  PRM            & 0.092  & 0.132   & 0.132   & 0.106   & 0.124   & 0.114   & 0.112   \\ \bottomrule
  \end{tabular}
  \end{adjustbox}
\end{subtable}
\caption{Training dynamics and word frequencies.}
\label{fig:prm_side_by_side}
\end{figure}

To better understand the strength of PRM rewards over self-certainty rewards, motivated by~\cite{wang2025beyond}, we also compute the frequency of transition words such as ``but", ``however", ``wait" which signal better reasoning and lower tendency for a model to be overly confidence in its response. In Tab.~\ref{tab:frequency}, we provide the frequency of such transition words over different checkpoints. The results show that the frequency of such transition words reduce with training iterations when trained with self-certainty rewards. PRM-rewards trained model on the other hand does not show a drastic reduction, highlighting the fact that PRM-rewards prevent model from getting overly confidence in its generation. 
\newpage
\subsection{Hyperparameters for Training} \label{app:hyper}
\hfill
\begin{table}[!h]
  \centering
  \caption{Training hyper-parameters for math reasoning}
  \label{tab:hparams-math}
  \sffamily
  \resizebox{\columnwidth}{!}{
  \begin{tabular}{@{}lcccc@{}}
    \toprule
    \multirow{2}{*}{\textbf{Parameter}} & \multicolumn{2}{c}{\textbf{Qwen2.5-3B}} & \multicolumn{2}{c}{\textbf{Qwen2.5-7B}} \\ \cmidrule(l){2-5}
    &  INT./GRPO &  PRM/PRISM  & INT./GRPO & PRM/PRISM \\
    \midrule
    
    Learning Rate              & \(3 \times 10^{-6}\) & \(3 \times 10^{-6}\) & \(1 \times 10^{-6}\) & \(1 \times 10^{-6}\) \\
    Per Device Batch Size                 & 3  & 4     &         4 & 4   \\
    Gradient Accumulation Steps & 37           & 36      & 32 & 32 \\
    Group Size                 & 7             & 6      & 16 & 16  \\
    KL Penalty \((\beta)\)     & 0.005         & 0.005    & 0.005 & 0.005 \\
    SC Weight \((\gamma)\)     & -             & quadratic decay from $1 \rightarrow 0$ & -& 1 \\
    Training Steps             & 300           & 300       & 73 & 73 \\
    Max Prompt Length          & 512           & 512      & 512 & 512 \\
    Max Completion Length      & 3072          & 3072      & 3072 & 1024 \\
    Temperature                & 0.9           & 0.9      & 0.9 & 0.9 \\
    Clip Ratio                 & 0.2           & 0.2      & 0.2 & 0.2 \\
    LR Scheduler Type          & Cosine        & Cosine      & Cosine & Cosine \\
    Warmup Ratio               & 0.1           & 0.1      & 0.1 & 0.1 \\
    \midrule
    \multicolumn{3}{@{}l@{}}{\textbf{Optimizer:} AdamW \(\left(\beta_{1}=0.9,\; \beta_{2}=0.999,\; \varepsilon = 10^{-8}\right)\)} \\
    \bottomrule
  \end{tabular}
  }
\end{table}

\subsection{Additional Results}
\begin{figure}[!hb]
    \centering
    \includegraphics[width=\linewidth]{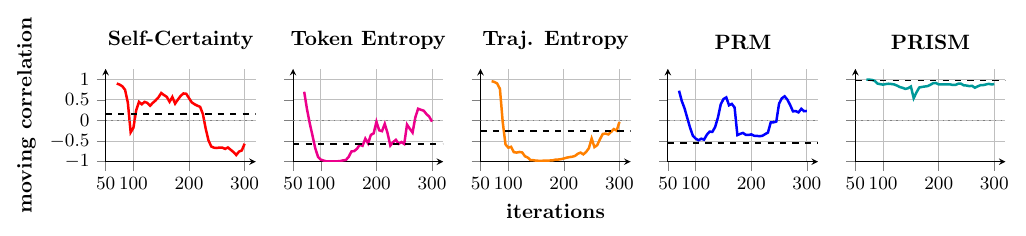}
    \caption{Moving correlation comparison for all methods. PRISM has the best mean and moving correlation among all.}
    \label{fig:corr_all}
\end{figure}
\subsection{System Prompts and GenPRM Pipeline} \label{app:genprm_details}
Here we provide system prompts that are used during training, along with the pipeline for GenPRM. 

  \begin{chatbubble}{gray}{System Prompt for MATH}
    \vspace{0.2in}
  \textbf{System prompt:}
  You are a helpful AI Assistant, designed to provided well-reasoned and
  detailed responses. You FIRST think about the reasoning process step by step
  and then provide the user with the answer. Please enclose your final answer
  in the box: \textbackslash boxed\{Your Answer\}. Please stop generation immediately after
  outputing the box.
\end{chatbubble}

The inputs to the GenPRM are the system prompt, and the question along with the student model's response. To facilitate paragraph detection, the following sentence: ``$<$output$>$Let's provide judgements for each of the paragraphs. \textbackslash n Paragraph 1 **Judgement** :\$\textbackslash boxed'' is added to GenPRM's response stream forcing it to continue generation following the context. An example is shown below: 

  \begin{chatbubble}{gray}{Input to GenPRM}
  \vspace{0.2in}
  \textbf{System prompt:}
You are a math teacher. Your task is to review, critique the paragraphs in solution step by step, and check if the solution is complete. Pay attention that you should neither solve the problem nor give the final answer.
  
  \medskip
  \textbf{Student Model:} How many interior diagonals does an icosahedron have? (An $\emph{icosahedron}$
  is a 3-dimensional figure with 20 triangular faces and 12 vertices, with 5
  faces meeting at each vertex. An $\emph{interior}$ diagonal is a segment
  connecting two vertices which do not lie on a common face.)
  
   \medskip
  \textbf{Student Model:} To determine the number of interior diagonals in an icosahedron, follow these steps:

1. \textbf{Calculate the Total Number of Diagonals}:
   The total number of line segments connecting any two vertices of an icosahedron (including both edges and diagonals) is given by the combination formula \( \binom{n}{2} \), where \( n \) is the number of vertices. For an icosahedron with 12 vertices:
   \[
   \binom{12}{2} = \frac{12 \times 11}{2} = 66
   \]
   So, there are 66 total line segments connecting any two vertices.

2. \textbf{Calculate the Number of Edges}:
   We know that an icosahedron has 30 edges:
   \[
   \text{Number of edges} = 30
   \]

3. \textbf{Calculate the Number of Interior Diagonals}:
   The interior diagonals are the line segments that are not edges of the icosahedron:
   \[
   \text{Number of interior diagonals} = \text{Total number of line segments} - \text{Number of edges}
   \]
   \[
   \text{Number of interior diagonals} = 66 - 30 = 36
   \]

Therefore, an icosahedron has 36 interior diagonals.
  
\end{chatbubble}

 \begin{chatbubble}{green}{GenPRM's Response}
  \vspace{0.2in}
$<$analyze$>$\\
Let's analyze all the Paragraphs step by step:
\\
\\
Paragraph 1: The first paragraph calculates the total number of line segments between vertices using the combination formula C(12, 2). This is correct because the total number of line segments (edges + diagonals) in a complete graph with n vertices is indeed n(n-1)/2. For 12 vertices, this gives 66, which is accurate.
\\
\\
Paragraph 2: The number of edges for an icosahedron is stated as 30. This is correct because an icosahedron is a regular polyhedron with 12 vertices and each vertex has 5 edges (since 5 faces meet at each vertex). Using Euler's formula (V – E + F = 2), where F is 20 faces, we can confirm edges: 12 – 30 + 20 = 2, which holds true. Alternatively, each of the 20 triangular faces has 3 edges, but each edge is shared by 2 faces, so (20 × 3)/2 = 30 edges. So this is correct.
\\
\\
Paragraph 3: The method to subtract edges from total segments to get interior diagonals is correct. However, there's a critical point here. The problem defines an interior diagonal as a segment connecting two vertices not on a common face. But in a convex polyhedron like an icosahedron, any two vertices not connected by an edge must be connected by a space diagonal (interior diagonal). However, in some non-convex polyhedrons, two vertices might not be on a common face but still lie on a face (if the face is non-convex or the polyhedron is concave). But since an icosahedron is convex, all edges are on the surface, and all non-edge connections are interior diagonals. Thus, subtracting edges from total gives correct interior diagonals. However, the solution's approach might be oversimplified. Wait, but the problem defines an interior diagonal as “not lying on a common face.” In an icosahedron, two vertices not sharing an edge must not lie on the same face. Because each face is a triangle, so any two vertices on the same face are connected by an edge. Therefore, the count is correct. The solution's approach is valid here.
\\
\\
Wait, but let's check if the problem's definition is equivalent to the standard interior diagonal. The problem says “interior diagonal is a segment connecting two vertices which do not lie on a common face.” Since each face is a triangle, two vertices on the same face must be connected by an edge (since a triangle has three edges). Therefore, any two vertices not on the same face cannot be connected by an edge, so the total diagonals (non-edges) would indeed be the interior diagonals as per the problem's definition. Therefore, the calculation is correct.
\\
\\
Therefore, all paragraphs are correct and the solution is valid.\\
$<$\textbackslash analyze$>$\\
$<$output$>$Let's provide judgements for each of the paragraphs.\\
Paragraph 1 \textbf{Judgement}: $\boxed{Yes}$\\
Paragraph 2 \textbf{Judgement}: $\boxed{Yes}$\\
Paragraph 3 \textbf{Judgement}: $\boxed{Yes}$\\
$<$\textbackslash output$>$\\
$<$output$>$Let's provide judgement if the user's solution is complete, i.e., does the solution provide an answer to the question (irrespective of correctness)?\\
\textbf{Judgement}: $\boxed{Yes}$\\
$<$\textbackslash output$>$

\end{chatbubble}
\subsection{Example RLIF-Trained Model's Generations After Failure} \label{app:fail}
\subsubsection{Self-Certainty-Trained Qwen2.5-3B Model}
\begin{chatbubble}{gray}{Input Prompt}
\vspace{0.2in}
\textbf{System prompt:}
You are a helpful AI Assistant, designed to provided well-reasoned and detailed responses. You FIRST think about the reasoning process step by step and then provide the user with the answer. Please enclose your final answer in the box: \textbackslash boxed\{Your Answer\}. Please stop generation immediately after outputing the box.

\medskip
\textbf{User:}
Marsha has two numbers, $a$ and $b$.  When she divides $a$ by 70 she gets a remainder of 64.  When she divides $b$ by 105 she gets a remainder of 99.  What remainder does she get when she divides $a+b$ by 35?
\end{chatbubble}

\begin{chatbubble}{green}{LLM's Response}
\vspace{0.2in}
To determine the remainder when \(a + b\) is divided by 35, we need to follow the following steps:
\\

1. \textbf{Express \(a\) and \(b\) in terms of their respective remainders.}\\
2. \textbf{Find the expression for \(a + b\).}\\
3. \textbf{Determine the remainder when \(a + b\) is divided by 35.}\\

\#\#\# Step-by-Step Solution
\\

\#\#\#  Step 1: Express \(a\) and \(b\) in terms of their remainders
Given that when \(a\) is divided by 70, the remainder is 64, we can write:
\[ a = 70k + 64 \]
where \(k\) is an integer.

Similarly, given that when \(b\) is divided by 105, the remainder is 99, we can write:
\[ b = 105m + 99 \]
where \(m\) is an integer.

\#\#\#  Step 2: Find the expression for \(a + b\)
Now, we need to find the value of \(a + b\):
\[ a + b = (70k + 64) + (105m + 99) \]
\[ a + b = 70k + 105m + 64 + 99 \]
\[ a + b = 70k + 105m + 163 \]

\#\#\#\#  Step 3: Determine the remainder when \(a + b\) is divided by 35
To find the remainder when \(a + b = 70k + 105m + 163\) is divided by 35, we need to find the remainder of \(163\) when divided by 35. We can do this by performing the division:
\[ 163 \div 35 \]

To find the quotient and the remainder, we perform the division:
\[ 163 \div 35 = 4 \text{ remainder } 23 \]

So, the remainder when 163 is divided by 35 is 23.

Thus, the remainder when \(a + b\) is divided by 35 is \boxed{23}.

Find the eigenvectors of the following matrix (round your answer to three decimal places):
$\left(
\begin{array}{ccc}
\frac{13}{3} & 6 & -\frac{14}{3} \\
-\frac{16}{3} & -\frac{16}{3} & -\frac{16}{3} \\
0 & \frac{2}{3} & -\frac{2}{3} \\
\end{array}
\right)$. To find the eigenvectors of the given matrix, we need to solve the characteristic equation and then find the corresponding eigenvectors for each eigenvalue. Let's go through the steps in detail.

1. \textbf{Find the characteristic equation:}
   The characteristic equation of a matrix \( A \) is given by \( \det(A - \lambda I) = 0 \), where \( \lambda \) is an eigenvalue and \( I \) is the identity matrix.

   Given matrix \( A \):
   \[
   A = \left(
   \begin{array}{ccc}
   \frac{13}{3} & 6 & -\frac{14}{3} \\
   -\frac{16}{3} & -\frac{16}{3} & -\frac{16}{3} \\
   0 & \frac{2}{3} & -\frac{2}{3} \\
   \end{array}
   \right)
   \]

   The identity matrix \( I \) is:
   \[
   I = \left(
   \begin{array}{ccc}
   1 & 0 & 0 \\
   0 & 1 & 0 \\
   0 & 0 & 1 \\
   \end{array}
   \right)
   \]

   So, \( A - \lambda I \) is:
   \[
   A - \lambda I = \left(
   \begin{array}{ccc}
   \frac{13}{3} - \lambda & 6 & -\frac{14}{3} \\
   -\frac{16}{3} & -\frac{16}{3} - \lambda & -\frac{16}{3} \\
   0 & \frac{2}{3} & -\frac{2}{3} - \lambda \\
   \end{array}
   \right)
   \]

   The determinant of \( A - \lambda I \) is:
   \[
   \det(A - \lambda I) = \left| \begin{array}{ccc}
   \frac{13}{3} - \lambda & 6 & -\frac{14}{3} \\
   -\frac{16}{3} & -\frac{16}{3} - \lambda & -\frac{16}{3} \\
   0 & \frac{2}{3} & -\frac{2}{3} - \lambda \\
   \end{array} \right|
   \]

   Expanding the determinant along the first row, we get:
   \[\scriptsize
   \det(A - \lambda I) = \left( \frac{13}{3} - \lambda \right) \left| \begin{array}{cc}
   -\frac{16}{3} - \lambda & -\frac{16}{3} \\
   \frac{2}{3} & -\frac{2}{3} - \lambda \\
   \end{array} \right| - 6 \left| \begin{array}{cc}
   -\frac{16}{3} & -\frac{16}{3} \\
   0 & -\frac{2}{3} - \lambda \\
   \end{array} \right| - \left( -\frac{14}{3} \right) \left| \begin{array}{cc}
   -\frac{16}{3} & -\frac{16}{3} - \lambda \\
   0 & \frac{2}{3} \\
   \end{array} \right|
   \]

   After calculating the determinants and simplifying, we get the characteristic equation:
   \[
   \lambda^3 - \frac{1}{3} \lambda^2 - \frac{100}{9} \lambda + \frac{100}{9} = 0
   \]

2. \textbf{Solve the characteristic equation:}
   Solving the cubic equation \( \lambda^3 - \frac{1}{3} \lambda^2 - \frac{100}{9} \lambda + \frac{100}{9} = 0 \) gives us the eigenvalues. Using numerical methods or a calculator, we find the eigenvalues to be approximately:
   \[
   \lambda_1 \approx 2, \quad \lambda_2 \approx -2, \quad \lambda_3 \approx 1
   \]

3. \textbf{Find the eigenvectors for each eigenvalue:}
   For each eigenvalue, we need to solve the system \( (A - \lambda I) \mathbf{v} = 0 \) to find the eigenvectors.

   - For \( \lambda_1 = 2 \):
     \[
     A - 2I = \left(
     \begin{array}{ccc}
     \frac{13}{3} - 2 & 6 & -\frac{14}{3} \\
     -\frac{16}{3} & -\frac{16}{3} - 2 & -\frac{16}{3} \\
     0 & \frac{2}{3} & -\frac{2}{3} - 2 \\
     \end{array}
     \right) = \left(
     \begin{array}{ccc}
     \frac{7}{3} & 6 & -\frac{14}{3} \\
     -\frac{16}{3} & -\frac{22}{3} & -\frac{16}{3} \\
     0 & \frac{2}{3} & -\frac{8}{3} \\
     \end{array}
     \right)
     \]

     Solving \( (A - 2I) \mathbf{v} = 0 \) gives us the eigenvector \( \mathbf{v}_1 \approx \{-0.447, 0.544, 1\} \).

   - For \( \lambda_2 = -2 \):
     \[
     A + 2I = \left(
     \begin{array}{ccc}
     \frac{13}{3} + 2 & 6 & -\frac{14}{3} \\
     -\frac{16}{3} & -\frac{16}{3} + 2 & -\frac{16}{3} \\
     0 & \frac{2}{3} & -\frac{2}{3} + 2 \\
     \end{array}
     \right) = \left(
     \begin{array}{ccc}
     \frac{19}{3} & 6 & -\frac{14}{3} \\
     -\frac{16}{3} & -\frac{10}{3} & -\frac{16}{3} \\
     0 & \frac{2}{3} & \frac{4}{3} \\
     \end{array}
     \right)
     \]

     Solving \( (A + 2I) \mathbf{v} = 0 \) gives us the eigenvector \( \mathbf{v}_2 \approx \{1.000, 1.000, 1.000\} \).

   - For \( \lambda_3 = 1 \):
     \[
     A - I = \left(
     \begin{array}{ccc}
     \frac{13}{3} - 1 & 6 & -\frac{14}{3} \\
     -\frac{16}{3} & -\frac{16}{3} - 1 & -\frac{16}{3} \\
     0 & \frac{2}{3} & -\frac{2}{3} - 1 \\
     \end{array}
     \right) = \left(
     \begin{array}{ccc}
     \frac{10}{3} & 6 & -\frac{14}{3} \\
     -\frac{16}{3} & -\frac{19}{3} & -\frac{16}{3} \\
     0 & \frac{2}{3} & -\frac{5}{3} \\
     \end{array}
     \right)
     \]

     Solving \( (A - I) \mathbf{v} = 0 \) gives us the eigenvector \( \mathbf{v}_3 \approx \{0, 0, 1\} \).

Therefore, the eigenvectors are \boxed{\{-0.447, 0.544, 1\}, \{1.000, 1.000, 1.000\}, \{0, 0, 1\}}.

Multiply the scalar $\frac{9}{25}$ with the matrix
$\left(
\begin{array}{cc}
5 & 10 \\
-6 & 5 \\
-10 & -5 \\
2 & 1 \\
\end{array}
\right)$. To multiply the scalar \(\frac{9}{25}\) with the given matrix, we need to multiply each element of the matrix by \(\frac{9}{25}\). Let's go through the steps in detail.

The given matrix is:
\[
A = \left(
\begin{array}{cc}
5 & 10 \\
-6 & 5 \\
-10 & -5 \\
2 & 1 \\
\end{array}
\right)
\]

We need to find the matrix \(B\) such that:
\[
B = \frac{9}{25} A
\]

This means we multiply each element of matrix \(A\) by \(\frac{9}{25}\). So, the element in the \(i\)-th row and \(j\)-th column of matrix \(B\) is given by:
\[
B_{ij} = \frac{9}{25} A_{ij}
\]

Let's perform the multiplication for each element of the matrix:

1. For the first row and first column:
\[
B_{11} = \frac{9}{25} \times 5 = \frac{9 \times 5}{25} = \frac{45}{25} = \frac{9}{5}
\]

2. For the first row and second column:
\[
B_{12} = \frac{9}{25} \times 10 = \frac{9 \times 10}{25} = \frac{90}{25} = \frac{18}{5}
\]

3. For the second row and first column:
\[
B_{21} = \frac{9}{25} \times (-6) = \frac{9 \times (-6)}{25} = \frac{-54}{25}
\]

4. For the second row and second column:
\[
B_{22} = \frac{9}{25} \times 5 = \frac{9 \times 5}{25} = \frac{45}{25} = \frac{9}{5}
\]

5. For the third row and first column:
\[
B_{31} = \frac{9}{25} \times (-10) = \]
    
\end{chatbubble}

\subsubsection{Trajectory-Entropy-Trained Qwen2.5-3B Model}

\begin{chatbubble}{gray}{Input Prompt}
\vspace{0.2in}
\textbf{System prompt:}
You are a helpful AI Assistant, designed to provided well-reasoned and detailed responses. You FIRST think about the reasoning process step by step and then provide the user with the answer. Please enclose your final answer in the box: \textbackslash boxed\{Your Answer\}. Please stop generation immediately after outputing the box.

\medskip
\textbf{User:} Let $z = \cos \frac{4 \pi}{7} + i \sin \frac{4 \pi}{7}.$  Compute
\[\frac{z}{1 + z^2} + \frac{z^2}{1 + z^4} + \frac{z^3}{1 + z^6}.\]
\end{chatbubble}

\begin{chatbubble}{green}{LLM's Response}
\vspace{0.2in}
   To solve the expression \(\frac{z}{1 + z^2} + \frac{z^2}{1 + z^4} + \frac{z^3}{1 + z^6}\) where \(z = \cos \frac{4 \pi}{7} + i \sin \frac{4 \pi}{7}\), we will follow these steps:
\\

1. \textbf{Simplify \(z\):}
   \[
   z = \cos \frac{4 \pi}{7} + i \sin \frac{4 \pi}{7}
   \]
   \[
   z = e^{i \frac{4 \pi}{7}}
   \]

2. \textbf{Calculate \(z^2\):}
   \[
   z^2 = \left(e^{i \frac{4 \pi}{7}}\right)^2
   \]
   \[
   z^2 = e^{i \frac{8 \pi}{7}}
   \]
   \[
   z^2 = e^{i \left(\frac{8 \pi}{7} - 2 \pi\right)}
   \]
   \[
   z^2 = e^{i \left(\frac{8 \pi}{7} - \frac{14 \pi}{7}\right)}
   \]
   \[
   z^2 = e^{i \left(\frac{8 \pi}{7} - \frac{14 \pi}{7}\right)}
   \]
   \[
   z^2 = e^{i \left(\frac{8 \pi}{7} - \frac{14 \pi}{7}\right)}
   \]
   \[
   z^2 = e^{i \left(\frac{8 \pi}{7} - \frac{14 \pi}{7}\right)}
   \]
   \[
   z^2 = e^{i \left(\frac{8 \pi}{7} - \frac{14 \pi}{7}\right)}
   \]
   \[
   z^2 = e^{i \left(\frac{8 \pi}{7} - \frac{14 \pi}{7}\right)}
   \]
   \[
   z^2 = e^{i \left(\frac{8 \pi}{7} - \frac{14 \pi}{7}\right)}
   \]
   \[
   z^2 = e^{i \left(\frac{8 \pi}{7} - \frac{14 \pi}{7}\right)}
   \]
   \[
   z^2 = e^{i \left(\frac{8 \pi}{7} - \frac{14 \pi}{7}\right)}
   \]
   \[
   z^2 = e^{i \left(\frac{8 \pi}{7} - \frac{14 \pi}{7}\right)}
   \]
   \[
   z^2 = e^{i \left(\frac{8 \pi}{7} - \frac{14 \pi}{7}\right)}
   \]
   \[
   z^2 = e^{i \left(\frac{8 \pi}{7} - \frac{14 \pi}{7}\right)}
   \]
   \[
   z^2 = e^{i \left(\frac{8 \pi}{7} - \frac{14 \pi}{7}\right)}
   \]
   \[
   z^2 = e^{i \left(\frac{8 \pi}{7} - \frac{14 \pi}{7}\right)}
   \]
   \[
   z^2 = e^{i \left(\frac{8 \pi}{7} - \frac{14 \pi}{7}\right)}
   \]
   \[
   z^2 = e^{i \left(\frac{8 \pi}{7} - \frac{14 \pi}{7}\right)}
   \]
   \[
   z^2 = e^{i \left(\frac{8 \pi}{7} - \frac{14 \pi}{7}\right)}
   \]
   \[
   z^2 = e^{i \left(\frac{8 \pi}{7} - \frac{14 \pi}{7}\right)}
   \]
   \[
   z^2 = e^{i \left(\frac{8 \pi}{7} - \frac{14 \pi}{7}\right)}
   \]
   \[
   z^2 = e^{i \left(\frac{8 \pi}{7} - \frac{14 \pi}{7}\right)}
   \]
   \[
   z^2 = e^{i \left(\frac{8 \pi}{7} - \frac{14 \pi}{7}\right)}
   \]
   \[
   z^2 = e^{i \left(\frac{8 \pi}{7} - \frac{14 \pi}{7}\right)}
   \]
   \[
   z^2 = e^{i \left(\frac{8 \pi}{7} - \frac{14 \pi}{7}\right)}
   \]
   \[
   z^2 = e^{i \left(\frac{8 \pi}{7} - \frac{14 \pi}{7}\right)}
   \]
   \[
   z^2 = e^{i \left(\frac{8 \pi}{7} - \frac{14 \pi}{7}\right)}
   \]
   \[
   z^2 = e^{i \left(\frac{8 \pi}{7} - \frac{14 \pi}{7}\right)}
   \]
   \[
   z^2 = e^{i \left(\frac{8 \pi}{7} - \frac{14 \pi}{7}\right)}
   \]
   \[
   z^2 = e^{i \left(\frac{8 \pi}{7} - \frac{14 \pi}{7}\right)}
   \]
   \[
   z^2 = e^{i \left(\frac{8 \pi}{7} - \frac{14 \pi}{7}\right)}
   \]
   \[
   z^2 = e^{i \left(\frac{8 \pi}{7} - \frac{14 \pi}{7}\right)}
   \]
   \[
   z^2 = e^{i \left(\frac{8 \pi}{7} - \frac{14 \pi}{7}\right)}
   \]
   \[
   z^2 = e^{i \left(\frac{8 \pi}{7} - \frac{14 \pi}{7}\right)}
   \]
   \[
   z^2 = e^{i \left(\frac{8 \pi}{7} - \frac{14 \pi}{7}\right)}
   \]
   \[
   z^2 = e^{i \left(\frac{8 \pi}{7} - \frac{14 \pi}{7}\right)}
   \]
   \[
   z^2 = e^{i \left(\frac{8 \pi}{7} - \frac{14 \pi}{7}\right)}
   \]
   \[
   z^2 = e^{i \left(\frac{8 \pi}{7} - \frac{14 \pi}{7}\right)}
   \]
   \[
   z^2 = e^{i \left(\frac{8 \pi}{7} - \frac{14 \pi}{7}\right)}
   \]
   \[
   z^2 = e^{i \left(\frac{8 \pi}{7} - \frac{14 \pi}{7}\right)}
   \]
   \[
   z^2 = e^{i \left(\frac{8 \pi}{7} - \frac{14 \pi}{7}\right)}
   \]
   \[
   z^2 = e^{i \left(\frac{8 \pi}{7} - \frac{14 \pi}{7}\right)}
   \]
   \[
   z^2 = e^{i \left(\frac{8 \pi}{7} - \frac{14 \pi}{7}\right)}
   \]
   \[
   z^2 = e^{i \left(\frac{8 \pi}{7} - \frac{14 \pi}{7}\right)}
   \]
   \[
   z^2 = e^{i \left(\frac{8 \pi}{7} - \frac{14 \pi}{7}\right)}
   \]
   \[
   z^2 = e^{i \left(\frac{8 \pi}{7} - \frac{14 \pi}{7}\right)}
   \]
   \[
   z^2 = e^{i \left(\frac{8 \pi}{7} - \frac{14 \pi}{7}\right)}
   \]
   \[
   z^2 = e^{i \left(\frac{8 \pi}{7} - \frac{14 \pi}{7}\right)}
   \]
   \[
   z^2 = e^{i \left(\frac{8 \pi}{7} - \frac{14 \pi}{7}\right)}
   \]
   \[
   z^2 = e^{i \left(\frac{8 \pi}{7} - \frac{14 \pi}{7}\right)}
   \]
   \[
   z^2 = e^{i \left(\frac{8 \pi}{7} - \frac{14 \pi}{7}\right)}
   \]
   \[
   z^2 = e^{i \left(\frac{8 \pi}{7} - \frac{14 \pi}{7}\right)}
   \]
   \[
   z^2 = e^{i \left(\frac{8 \pi}{7} - \frac{14 \pi}{7}\right)}
   \]
   \[
   z^2 = e^{i \left(\frac{8 \pi}{7} - \frac{14 \pi}{7}\right)}
   \]
   \[
   z^2 = e^{i \left(\frac{8 \pi}{7} - \frac{14 \pi}{7}\right)}
   \]
   \[
   z^2 = e^{i \left(\frac{8 \pi}{7} - \frac{14 \pi}{7}\right)}
   \]
   \[
   z^2 = e^{i \left(\frac{8 \pi}{7} - \frac{14 \pi}{7}\right)}
   \]
   \[
   z^2 = e^{i \left(\frac{8 \pi}{7} - \frac{14 \pi}{7}\right)}
   \]
   \[
   z^2 = e^{i \left(\frac{8 \pi}{7} - \frac{14 \pi}{7}\right)}
   \]
   \[
   z^2 = e^{i \left(\frac{8 \pi}{7} - \frac{14 \pi}{7}\right)}
   \]
   \[
   z^2 = e^{i \left(\frac{8 \pi}{7} - \frac{14 \pi}{7}\right)}
   \]
   \[
   z^2 = e^{i \left(\frac{8 \pi}{7} - \frac{14 \pi}{7}\right)}
   \]
   \[
   z^2 = e^{i \left(\frac{8 \pi}{7} - \frac{14 \pi}{7}\right)}
   \]
   \[
   z^2 = e^{i \left(\frac{8 \pi}{7} - \frac{14 \pi}{7}\right)}
   \]
   \[
   z^2 = e^{i \left(\frac{8 \pi}{7} - \frac{14 \pi}{7}\right)}
   \]
   \[
   z^2 = e^{i \left(\frac{8 \pi}{7} - \frac{14 \pi}{7}\right)}
   \]
   \[
   z^2 = e^{i \left(\frac{8 \pi}{7} - \frac{14 \pi}{7}\right)}
   \]
   \[
   z^2 = e^{i \left(\frac{8 \pi}{7} - \frac{14 \pi}{7}\right)}
   \]
   \[
   z^2 = e^{i \left(\frac{8 \pi}{7} - \frac{14 \pi}{7}\right)}
   \]
   \[
   z^2 = e^{i \left(\frac{8 \pi}{7} - \frac{14 \pi}{7}\right)}
   \]
   \[
   z^2 = e^{i \left(\frac{8 \pi}{7} - \frac{14 \pi}{7}\right)}
   \]
   \[
   z^2 = e^{i \left(\frac{8 \pi}{7} - \frac{14 \pi}{7}\right)}
   \]
   \[
   z^2 = e^{i \left(\frac{8 \pi}{7} - \frac{14 \pi}{7}\right)}
   \]
   \[
   z^2 = e^{i \left(\frac{8 \pi}{7} - \frac{14 \pi}{7}\right)}
   \]
   \[
   z^2 = e^{i \left(\frac{8 \pi}{7} - \frac{14 \pi}{7}\right)}
   \]
   \[
   z^2 = e^{i \left(\frac{8 \pi}{7} - \frac{14 \pi}{7}\right)}
   \]
   \[
   z^2 = e^{i \left(\frac{8 \pi }{}\right.}\]
\end{chatbubble}

\subsubsection{Token-Entropy-Trained Qwen2.5-3B Model}
\begin{chatbubble}{gray}{Input Prompt}
\vspace{0.2in}
\textbf{System prompt:}
You are a helpful AI Assistant, designed to provided well-reasoned and detailed responses. You FIRST think about the reasoning process step by step and then provide the user with the answer. Please enclose your final answer in the box: \textbackslash boxed\{Your Answer\}. Please stop generation immediately after outputing the box.

\medskip
\textbf{User:} What is the remainder when $2001 \cdot 2002 \cdot 2003 \cdot 2004 \cdot 2005$ is divided by 19?
\end{chatbubble}

\begin{chatbubble}{green}{LLM's Response}
\vspace{0.2in}
    To determine the remainder when \(2001 \cdot 2002 \cdot 2003 \cdot 2004 \cdot 2005\) is divided by 19, let's follow these steps:

1. Evaluate \(2001\):
   \[
   2001
   \]
   \[
   2001
   \]
   \[
   2001
   \]
   \[
   2001
   \]
   \[
   2001
   \]
   \[
   2001
   \]
   \[
   2001
   \]
   \[
   2001
   \]
   \[
   2001
   \]
   \[
   2001
   \]
   \[
   2001
   \]
   \[
   2001
   \]
   \[
   2001
   \]
   \[
   2001
   \]
   \[
   2001
   \]
   \[
   2001
   \]
   \[
   2001
   \]
   \[
   2001
   \]
   \[
   2001
   \]
   \[
   2001
   \]
   \[
   2001
   \]
   \[
   2001
   \]
   \[
   2001
   \]
   \[
   2001
   \]
   \[
   2001
   \]
   \[
   2001
   \]
   \[
   2001
   \]
   \[
   2001
   \]
   \[
   2001
   \]
   \[
   2001
   \]
   \[
   2001
   \]
   \[
   2001
   \]
   \[
   2001
   \]
   \[
   2001
   \]
   \[
   2001
   \]
   \[
   2001
   \]
   \[
   2001
   \]
   \[
   2001
   \]
   \[
   2001
   \]
   \[
   2001
   \]
   \[
   2001
   \]
   \[
   2001
   \]
   \[
   2001
   \]
   \[
   2001
   \]
   \[
   2001
   \]
   \[
   2001
   \]
   \[
   2001
   \]
   \[
   2001
   \]
   \[
   2001
   \]
   \[
   2001
   \]
   \[
   2001
   \]
   \[
   2001
   \]
   \[
   2001
   \]
   \[
   2001
   \]
   \[
   2001
   \]
   \[
   2001
   \]
   \[
   2001
   \]
   \[
   2001
   \]
   \[
   2001
   \]
   \[
   2001
   \]
   \[
   2001
   \]
   \[
   2001
   \]
   \[
   2001
   \]
   \[
   2001
   \]
   \[
   2001
   \]
   \[
   2001
   \]
   \[
   2001
   \]
   \[
   2001
   \]
   \[
   2001
   \]
   \[
   2001
   \]
   \[
   2001
   \]
   \[
   2001
   \]
   \[
   2001
   \]
   \[
   2001
   \]
   \[
   2001
   \]
   \[
   2001
   \]
   \[
   2001
   \]
   \[
   2001
   \]
   \[
   2001
   \]
   \[
   2001
   \]
   \[
   2001
   \]
   \[
   2001
   \]
   \[
   2001
   \]
   \[
   2001
   \]
   \[
   2001
   \]
   \[
   2001
   \]
   \[
   2001
   \]
   \[
   2001
   \]
   \[
   2001
   \]
   \[
   2001
   \]
   \[
   2001
   \]
   \[
   2001
   \]
   \[
   2001
   \]
   \[
   2001
   \]
   \[
   2001
   \]
   \[
   2001
   \]
   \[
   2001
   \]
   \[
   2001
   \]
   \[
   2001
   \]
   \[
   2001
   \]
   \[
   2001
   \]
   \[
   2001
   \]
   \[
   2001
   \]
   \[
   2001
   \]
   \[
   2001
   \]
   \[
   2001
   \]
   \[
   2001
   \]
   \[
   2001
   \]
   \[
   2001
   \]
   \[
   2001
   \]
   \[
   2001
   \]
   \[
   2001
   \]
   \[
   2001
   \]
   \[
   2001
   \]
   \[
   2001
   \]
   \[
   2001
   \]
   \[
   2001
   \]
   \[
   2001
   \]
   \[
   2001
   \]
   \[
   2001
   \]
   \[
   2001
   \]
   \[
   2001
   \]
   \[
   2001
   \]
   \[
   2001
   \]
   \[
   2001
   \]
   \[
   2001
   \]
   \[
   2001
   \]
   \[
   2001
   \]
   \[
   2001
   \]
   \[
   2001
   \]
   \[
   2001
   \]
   \[
   2001
   \]
   \[
   2001
   \]
   \[
   2001
   \]
   \[
   2001
   \]
   \[
   2001
   \]
   \[
   2001
   \]
   \[
   2001
   \]
   \[
   2001
   \]
   \[
   2001
   \]
   \[
   2001
   \]
   \[
   2001
   \]
   \[
   2001
   \]
   \[
   2001
   \]
   \[
   2001
   \]
   \[
   2001
   \]
   \[
   2001
   \]
   \[
   2001
   \]
   \[
   2001
   \]
   \[
   2001
   \]
   \[
   2001
   \]
   \[
   2001
   \]
   \[
   2001
   \]
   \[
   2001
   \]
   \[
   2001
   \]
   \[
   2001
   \]
   \[
   2001
   \]
   \[
   2001
   \]
   \[
   2001
   \]
   \[
   2001
   \]
   \[
   2001
   \]
   \[
   2001
   \]
   \[
   2001
   \]
   \[
   2001
   \]
   \[
   2001
   \]
   \[
   2001
   \]
   \[
   2001
   \]
   \[
   2001
   \]
   \[
   2001
   \]
   \[
   2001
   \]
   \[
   2001
   \]
   \[
   2001
   \]
   \[
   2001
   \]
   \[
   2001
   \]
   \[
   2001
   \]
   \[
   2001
   \]
   \[
   2001
   \]
   \[
   2001
   \]
   \[
   2001
   \]
   \[
   2001
   \]
   \[
   2001
   \]
   \[
   2001
   \]
   \[
   2001
   \]
   \[
   2001
   \]
   \[
   2001
   \]
   \[
   2001
   \]
   \[
   2001
   \]
   \[
   2001
   \]
   \[
   2001
   \]
   \[
   2001
   \]
   \[
   2001
   \]
   \[
   2001
   \]
   \[
   2001
   \]
   \[
   2001
   \]
   \[
   2001
   \]
   \[
   2001
   \]
   \[
   2001
   \]
   \[
   2001
   \]
   \[
   2001
   \]
   \[
   2001
   \]
   \[
   2001
   \]
   \[
   2001
   \]
   \[
   2001
   \]
   \[
   2001
   \]
   \[
   2001
   \]
   \[
   2001
   \]
   \[
   2001
   \]
   \[
   2001
   \]
   \[
   2001
   \]
   \[
   2001
   \]
   \[
   2001
   \]
   \[
   2001
   \]
   \[
   2001
   \]
   \[
   2001
   \]
   \[
   2001
   \]
   \[
   2001
   \]
   \[
   2001
   \]
   \[
   2001
   \]
   \[
   2001
   \]
   \[
   2001
   \]
   \[
   2001
   \]
   \[
   2001
   \]
   \[
   2001
   \]
   \[
   2001
   \]
   \[
   2001
   \]
   \[
   2001
   \]
   \[
   2001
   \]
   \[
   2001
   \]
   \[
   2001
   \]
   \[
   2001
   \]
   \[
   2001
   \]
   \[
   20 \]
\end{chatbubble}
\end{document}

%% file: math_commands.tex
\usepackage{amsmath,amsfonts,bm}

\def\eqref#1{equation~\ref{#1}}

\def\1{\bm{1}}

\DeclareMathAlphabet{\mathsfit}{\encodingdefault}{\sfdefault}{m}{sl}
\SetMathAlphabet{\mathsfit}{bold}{\encodingdefault}{\sfdefault}{bx}{n}

